\documentclass[letterpaper, 10 pt, journal, twoside]{IEEEtran}

\IEEEoverridecommandlockouts                              

\usepackage{graphicx}
\usepackage{amsmath}
\usepackage{amssymb}
\usepackage{booktabs}
\usepackage{caption}
\usepackage{multirow}
\usepackage{rotating}
\usepackage{siunitx}
\usepackage{caption}
\usepackage{subcaption}
\usepackage[ruled,vlined,linesnumbered]{algorithm2e}
\usepackage{xcolor}
\usepackage{cuted}

\usepackage[inline]{enumitem}
\usepackage{xcolor}
\usepackage{microtype}
\usepackage{bbm}
\usepackage{dirtytalk}
\usepackage{cite}
\usepackage{dsfont}
\usepackage{fancyhdr}

\usepackage{hyperref}

\usepackage[acronym]{glossaries}
\newacronym{rpn}{RPN}{Region Proposal Network}
\newacronym{roi}{RoI}{Region of Interest}
\newacronym{iou}{IoU}{intersection-over-union}
\newacronym{nms}{NMS}{non-maximum suppression}
\newacronym{rgcn}{R-GCNs}{Relational Graph Convolutional Networks~\cite{Schlichtkrull2018}}
\newacronym{fpn}{FPN}{Feature Proposal Network}
\newacronym{kern}{KERN}{Knowledge-Embedded Routing Network \cite{Chen2019}}
\newacronym{bert}{BERT}{Bidirectional Encoder Representations from Transformers~\cite{Devlin2019}}
\newacronym{gru}{GRU}{Gated Recurrent Cells~\cite{Li2016}}
\newacronym{fcn}{FCN}{fully-connected network}
\newacronym{rcgn}{RCGN}{Relational Graph Neural Network}
\newacronym{ggnn}{GGNN}{Gated-Graph Neural Network}
\newacronym{mlp}{MLP}{Multi-Layer Perceptron}
\newacronym{coco}{COCO 2017}{Common Objects in Context~\cite{lin2014microsoft}}
\newacronym{lvis}{LVIS v1}{LVIS v1~\cite{gupta2019lvis}}
\newacronym{svg}{SVG}{singular value decomposition}
\newacronym{kge}{KGE}{knowledge graph embedded}
\newacronym{sota}{SoTA}{state-of-the-art}
\newacronym{mlr}{MLR}{multi logistic regression}
\newacronym{vos}{VOS}{visualizing similarities between objects}
\newacronym{bdd}{BDD100k}{Berkeley Deep Drive~\cite{yu2020bdd100k}}
\newacronym{mot}{MOT}{Multi-Object Tracking}
\newacronym{ssmot}{SSMOT}{Self-Supervised Multi-Object Tracking}
\newacronym{mot17}{MOT17}{~\cite{MOT16}}
\newacronym{ssl}{SSL}{self-supervised learning}
\newacronym{fps}{FPS}{frames per second}
\newacronym{msa}{MSA}{minimum-sum assigment}
\newacronym{tempo}{TempO}{temporal ordering pretext}
\newacronym{nn}{NN}{nearest-neighbor}
\newacronym{ucf}{UCF101}{UCF101~\cite{soomro2012ucf101}}
\newacronym{hota}{HOTA}{higher-order tracking accuracy~\cite{luiten2020hota}}
\newacronym{reid}{ReID}{re-identification}
\newacronym{vit}{ViT}{vision transformer~\cite{dosovitskiy2020image}}
\newacronym{mota}{MOTA}{multi-object tracking accuracy~\cite{bernardin2008clear}}
\newacronym{subco}{SubCo}{subsample consistency}
\newacronym{rnn}{RNN}{recurrent neural network}
\glsdisablehyper

\usepackage[capitalize]{cleveref}
\crefname{section}{Sec.}{Secs.}
\Crefname{section}{Section}{Sections}
\Crefname{table}{Table}{Tables}
\crefname{table}{Tab.}{Tabs.}
\crefname{algorithm}{Algo.}{Algos.}

\newcommand{\figref}[1]{Figure~\ref{#1}}
\newcommand{\tabref}[1]{Table~\ref{#1}}

\newcommand{\algref}[1]{Algorithm~\ref{#1}}
\newcommand{\equref}[1]{Equation~(\ref{#1})}

\usepackage{xspace}

\def\ie{\emph{i.e}.}

\renewcommand{\baselinestretch}{0.985}

\title{\LARGE \bf
Self-Supervised Multi-Object Tracking For Autonomous Driving \\From Consistency Across Timescales
}

\author{Christopher Lang$^{1,2}$, Alexander Braun$^{2}$, Lars Schillingmann$^{2}$, Abhinav Valada$^{1}$
    \thanks{$^{1}$University of Freiburg, $^{2}$Robert Bosch GmbH}%
}

\begin{document}

\maketitle
\thispagestyle{empty}
\pagestyle{empty}

\begin{abstract}
Self-supervised multi-object trackers have tremendous potential as they enable learning from raw domain-specific data. 
However, their re-identification accuracy still falls short compared to their supervised counterparts.
We hypothesize that this drawback results from formulating self-supervised objectives that are limited to single frames or frame pairs. 
Such formulations do not capture sufficient visual appearance variations to facilitate learning consistent re-identification features for autonomous driving when the frame rate is low, or object dynamics are high. 
In this work, we propose a training objective that enables self-supervised learning of re-identification features from multiple sequential frames by enforcing consistent association scores across short and long timescales. 
We perform extensive evaluations that demonstrate that re-identification features trained from longer sequences significantly reduce ID switches on standard autonomous driving datasets compared to existing self-supervised learning methods, which are limited to training on frame pairs. Using our proposed SubCo loss function, we set the new state of the art among self-supervised methods and even perform on par with fully supervised learning methods. 
\end{abstract}


\section{Introduction}
For any autonomous vehicle, it is crucial to accurately track and predict the movements of surrounding agents for safe and collision-free navigation. This task is particularly challenging in dense urban environments due to occlusions, illumination variation, and non-rigid motions, which require extensive data annotations that must often be updated as sensor setup or application areas change.
Self-supervised approaches offer a promising alternative, as they enable learning directly from raw sensor data recorded from the robotic platform rather than relying on human-annotated labels~\cite{lang2023self}.

To re-detect an object after occlusions or missed detections, multi-object trackers depend on visual features~\cite{wojke2017simple, bergmann2019tracking} to extract consistent and discriminative representations for each object throughout a sequence. 
Typical self-supervised learning methods are either trained with pseudo-identities from image augmentations~\cite{chen2020improved,caron2020swav,kim2022sslmot} and motion-based trackers~\cite{karthik2020simple,valverde2021there} or define identify-free pretext tasks that enforce a cyclic consistency constraint~\cite{wang2020cycas,chung2022ssat,liu2022utrack}.
However, these methods still suffer from numerous identity switches compared to supervised learning methods on the MOT17~\cite{MOT16} benchmark. We argue that such approaches suffer in handling appearance variations or occlusion scenarios, as these short-range comparisons and image augmentations only partially reflect real-world variances. These effects are more pronounced in autonomous driving settings with lower frame rates.

\begin{figure}
    \centering
    \includegraphics[width=1.0\linewidth]{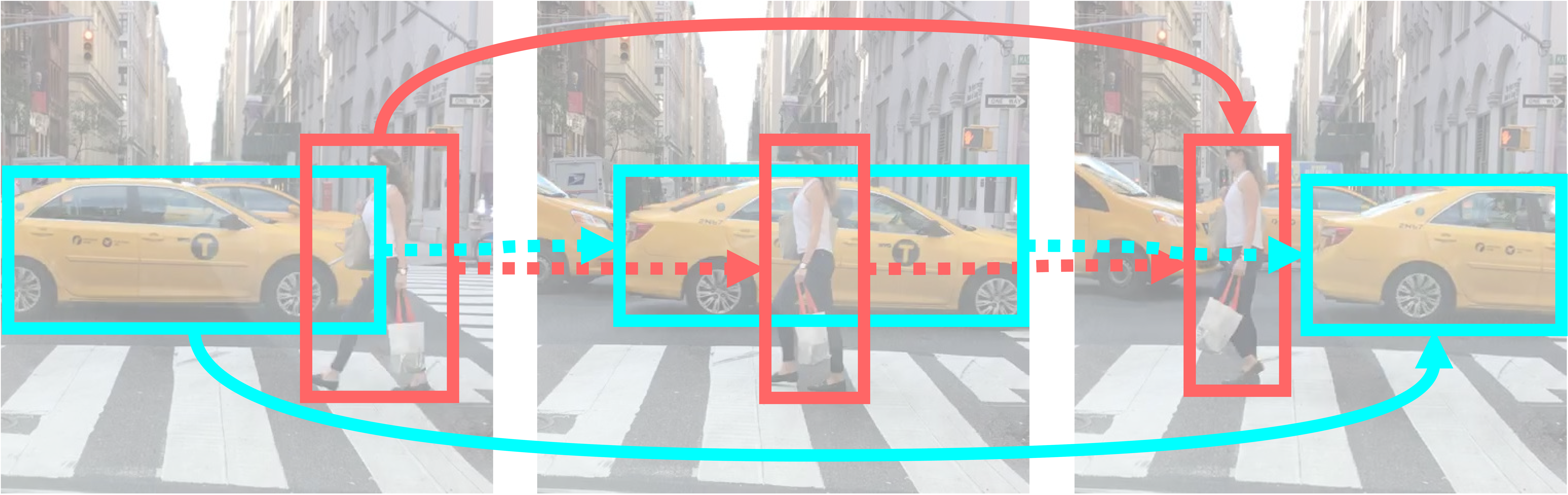}
    \caption{We propose the self-supervised tracking objective \gls{subco}, that learns \gls{reid} features for tracking object instances along a sequence by enforcing consistent association scores when tracking at short timescales (dotted lines) and long timescales (solid lines).}
    \label{fig:sequence_example}
    \vspace{-0.5cm}
\end{figure}

In this work, we propose to train associations across multiple frames instead of limiting them to frame pairs, from which we enable the \gls{reid} models to learn more robust and consistent feature representations. 
We formulate \gls{subco}, a self-supervised training objective that propagates association scores along a sequence of frames and enforces consistency between the propagated association scores and the scores directly computed between the start and end frame, as depicted in \figref{fig:sequence_example}. 
Our formulation incorporates the handling of track birth and track deletions to represent full occlusion scenarios in the loss computation, which allows us to train on unfiltered raw sequences.
In doing so, we enable the \gls{reid} model to learn temporally consistent features and handle appearance changes and partial occlusions occurring naturally in the training data.

We perform extensive evaluations on the \gls{bdd} benchmarks as well as multi-pedestrian tracking~\cite{MOT16} using the YOLO-X~\cite{yolox2021} detector and ablate over varying sequence lengths as well as loss terms to quantify the influence of our proposed loss formulation.
Our results show that the \gls{reid} features learned from this objective achieve the lowest number of ID switches compared to all the self-supervised baselines and reduce the overall number of ID switches by $22\%$ compared to the state-of-the-art ByteTrack~\cite{zhang2022bytetrack} architecture on the \gls{bdd} benchmark.
These results can be attributed to the ability of our proposed method to train on longer timescales, as we found that increasing sequence lengths consistently improved the association accuracy for both driving and pedestrian tracking datasets.
More importantly, our approach sets the new state of the art for self-supervised multi-object tracking and even performs on par with supervised tracking methods by opening up new avenues by exploiting long-term dependencies for self-supervised learning.

\section{Related Work}
We propose a self-supervised \acrshort{mot} method that builds on a pre-trained object detector and relies on \acrshort{mot} architectures.

{\parskip=3pt
\noindent\textit{\acrfull{mot}}: Multi-object trackers predict object locations and their instance identity in sequences.
Tracking-by-detection approaches typically use a two-stage process, first detecting objects in each frame and then linking detections across frames to form tracks.
Such methods have gained popularity in recent years due to the advancements in object detection algorithms, such as Faster R-CNN~\cite{bewley2016simple,wojke2017simple,pang2021quasi}, YOLO~\cite{zhang2022bytetrack,bastani2021visualspatial}, or CenterNet~\cite{wang2020cycas,zhou2020centertrack,chung2022ssat,kim2022sslmot}.}

Location and motion-based trackers~\cite{bewley2016simple,bergmann2019tracking} employ a Kalman Filter to propagate tracked instances and associate them among the high-scoring detection boxes using bipartite matching to maximize the overall \gls{iou}.
BYTE~\cite{zhang2022bytetrack} proposes to associate the low score detection boxes as well since the low confidence might stem from partial occlusions of a tracked object.
However, with increasing occlusion ratios, these methods suffer from high position uncertainty resulting in many lost tracks in busy environments or front-view cameras in autonomous driving.
Visual features of the tracked instances are therefore used to re-detect objects from their appearance~\cite{zeng2021motr,wojke2017simple,zhang2021fairmot,pang2021quasi,cai2022memot,fong2022panoptic}.
Such \gls{reid} features aim to generalize across many instances of an object type in varying illumination and occlusion cases. 
Therefore, it is critical to include such examples in the training of the tracking model.\looseness=-1

In contrast, multitask methods train object detection and tracking networks jointly. This is achieved by extracting \gls{reid} features from a shared backbone~\cite{zhang2021fairmot} and extending the bounding box prediction head with a single object tracking capability to regress the object location from the previous frame to the current frame~\cite{bergmann2019tracking,zhou2020centertrack}. However, the best-performing methods on MOT benchmarks, and especially \gls{ssl} methods, treat object detection and \gls{reid} feature estimation as two separate tasks, presumably since they can tailor the network architecture to each task and handle scale variations in the \gls{reid} task by rescaling the image crops used for extracting tracking features~\cite{zhang2021fairmot}. Nevertheless, the self-supervised tracking setting requires a pre-trained object detector.\looseness=-1

Approaches using a single network for object detection and tracking bounding boxes, such as CenterTrack~\cite{zhou2020centertrack}, either work offline or lack the ability to reconnect long-range tracks.
The latter scenarios require more advanced tracking logic accompanied by memory.
With the rise of query-based object detectors, such as DETR~\cite{zhu2020deformable}, the tracking by query propagation paradigm~\cite{zeng2021motr,meinhardt2022trackformer} has emerged. 
These methods learn to track queries to recall the same instance across different frames in online or offline settings.
However, there are no identity-free approaches to train such models in the literature, which we could compare against our proposed method.\looseness=-1

{\parskip=3pt
\noindent\textit{Self-supervised \acrshort{mot}}: \gls{ssmot} methods learn the association of object proposals without annotated track identities.
This includes the non-learnable trackers based on bounding box coordinates reviewed above and a variety of re-identification feature learning methods~\cite{wu2020tracklet}.}
Pseudo identity approaches generate labels from \gls{iou}-based, parameter-less, tracking approaches, clustering~\cite{wu2020tracklet}, or image augmentations that ’’hallucinate” tracking~\cite{zhou2020centertrack}. 
Also, self-supervised learning approaches from single images are used, as their representations are trained to be invariant augmentations~\cite{caron2020swav,chen2021mocov3,chen2020improved,wei2022maskedfeat} that simulate the temporal variances of detection box crops.

Such augmentation-based methods result in comparable high numbers of lost tracks, as they never see the object under motion, occlusion, or varying perspectives.
Such rectangular crops also include background regions of the original image. Consequently, they neglect to replicate the foreground-background interaction in real-world sequences that might contain important information about the motion or interaction of various tracks.

On the other hand, identity-free approaches derive hard and weak supervision signals from the raw data stream.
Supervision signals from color propagation~\cite{larsson2017colorization}, cycle-consistency~\cite{wang2019learning}, or optical flow~\cite{bochinski2018opticalflow} are proposed to train single object tracking models. However, Bastani \textit{et al.}~\cite{bastani2021visualspatial} argue that such approaches do not transfer to the multi-object setting, as the single-trackers tend to match new detections to tracks with the closest initial detections~\cite{bastani2021visualspatial}.
Annotation-free multi-object tracking strategies were therefore developed with tracking multiple regions, for instance, random patches~\cite{wang2019unsupervised} or detection boxes~\cite{wang2020cycas}, along cyclic associations.
However, such methods can never guarantee that all object instances are visible throughout the sequence. Consequently, they soften their association constraints and only train on image pairs.
We propose to track along multiple timescales to avoid backward-tracking in time. Consequently, we can lift this constraint of all objects being visible throughout the sequence. 
Bastani~\textit{et~al.}~\cite{bastani2021visualspatial} employ an input hiding scheme to simulate occlusions during training actively. 
They learn \gls{reid} features with an RNN-based tracker, with the learning objective to output similar transition matrices for distinct masking variations of the detections.

Our proposed \gls{subco} training task also derives the loss value from similarities of assignment matrices. 
However, our formulation is free of data augmentations, such as the input hiding scheme in~\cite{bastani2021visualspatial}, which allows our training to include the complete data variability in the raw sequence, for instance, partial occlusions, illumination variations, or non-rigid motions.
Additionally, our training strategy can be applied to various \gls{reid} model architectures and is not limited to \gls{rnn} models.

\section{Technical Approach}
\begin{figure*}
    \centering
    \includegraphics[width=0.9\textwidth]{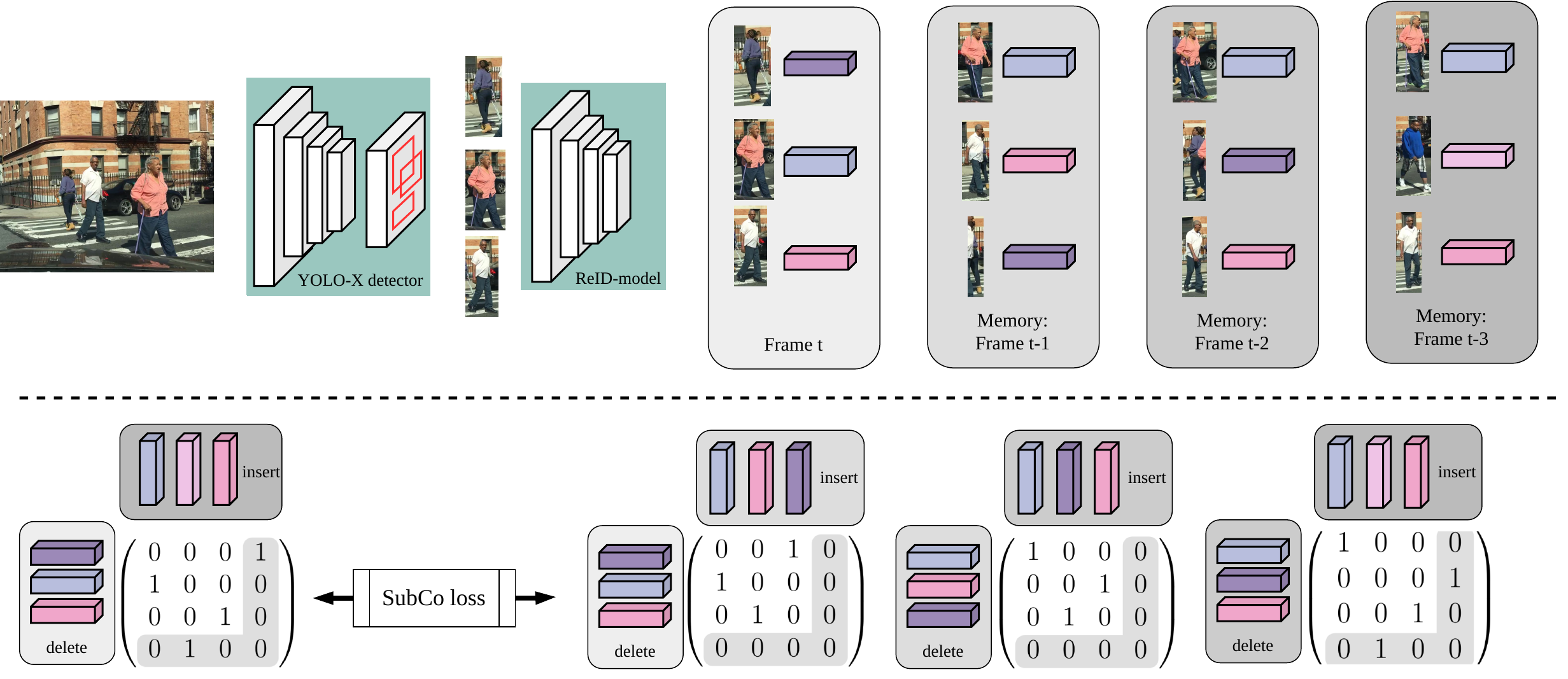}
    \caption{Illustration of the proposed training pipeline. 
    A YOLO-X~\cite{yolox2021} object detector with frozen parameters predicts a set of proposal boxes per frame. 
    Each proposal box is encoded into a ReID feature vector by cropping the respective image region and feeding it through a feature extractor as a ReID model.
    The bottom part depicts the computation of the \gls{subco} loss following \algref{algo:compute_loss}.
    The \gls{subco} loss is computed over a sequence of frame-wise ReID feature vectors by enforcing consistency between the association scores between the start and end frame (long-timescale) as well as pairwise along the sequence (short-timescale), computed following \algref{algo:compute_associations}.
    The association score matrices of the pairwise comparisons are accumulated by matrix multiplication, and a deletion vector is propagated along the sequence to keep track of lost tracks.
    The consistency loss is computed for association scores over all the not lost tracks.
    }
    \label{fig:system_diagram}
    \vspace{-0.2cm}
\end{figure*}

The objective in \gls{mot} is to predict multiple tracks in a video sequence $\mathbf{V} = \{ \mathbf{I}^t \}, \; t = 1, \dots, N$.
Each track in this video is represented as a tuple of a track identity $i$, $L_i$ tracking bounding boxes, a class label, as well as a \gls{reid} feature vector $\mathbf{t}_i \in \mathbb{R}^D$. 

The overall system architecture is depicted in \figref{fig:system_diagram}. 
The set of active tracks $\mathbf{T}^{t-1}$ up to time $t-1$ is updated online in a tracking-by-detection fashion, \ie  given the frame $\mathbf{I}^t$, a pre-trained object detector predicts a set of $K^t$ detection bounding boxes $\mathbf{P}^t = \{ \Vec{\mathbf{p}_{1}}, \dots, \Vec{\mathbf{p}}_{K^t} \}_n = \Theta (\mathbf{I}^t)$ where each detection $\mathbf{p}_{k}$ is represented by bounding box coordinates $\Vec{\mathbf{b}}_{k} \in \mathbb{R}^4$ and a class label $c_k$. 
Next, these detection boxes $\mathbf{P}^t$ are assigned to active tracks in $\mathbf{T}^{t-1}$, or used to initialize new tracks.
During tracking inference, the assignment is chosen to maximize the sum of pairwise association scores between the active tracks $\mathbf{T}^{t-1}$ and the detection boxes $\mathbf{P}^t$ in a one-to-one matching.
Following related works in self-supervised tracking~\cite{chung2022ssat,kim2022sslmot,wang2020cycas,liu2022utrack}, we derive the association scores from \glsfirst{reid} feature similarities.
Those \gls{reid} features are computed as a mapping of each bounding box region in the image into an \gls{reid} embedding space by $\mathbf{X}^t = \Phi(\mathbf{I}^t, \mathbf{P}^t )$ where $\mathbf{X}^t = \left[ \Vec{\mathbf{x}}_1, \dots, \Vec{\mathbf{x}}_{K^t}  \right] \in \mathbb{R}^{K^t \times D}$ is an embedding matrix composed of $K^t$ embedding vectors of dimension $D$.


The association score matrix is computed as the outer matrix product between $\mathbf{Y}^{t-1}$, that concatenates the \gls{reid} features of all active track feature $\Vec{\mathbf{t}}_i$ in $\mathbf{T}^{t-1}$ as column vectors, and the \gls{reid} features of the detection boxes $\mathbf{X}^t$ :
\begin{align}
    {}^{t-1}_{}\mathbf{S}^{}_{t} = \mathbf{Y}^{t-1} \left( \mathbf{X}^t\right)^T \in \mathbb{R}^{|\mathbf{T}^{t-1}| \times K^t}
    \label{eq:association_scores}
\end{align}
where $|\mathbf{T}^{t-1}|$ denotes the cardinality of the set $\mathbf{T}^{t-1}$.
In this work, the parameters of the \gls{reid} model $\Phi$ are trained using our proposed self-supervised \gls{subco} task.

\subsection{Self-Supervised ReID Feature Learning}

Our proposed \gls{subco} loss function is defined on the assignment matrices between frame pairs.
Ideally, the assignment matrix takes the form of a binary matrix that describes a bipartite matching graph between the set of tracks $\mathbf{T}$ with the set of detections $\mathbf{P}$ as:
\begin{align}
    \mathbf{A} \in \{0,1\}^{(|\mathbf{T}|) \times (|\mathbf{P}| + |\mathbf{D}|)}, \text{s.t.} \sum_j [\mathbf{A}]_{i,j} \leq 1 \forall i \;,
\end{align}
where $[\mathbf{A}]_{ij}=1$ indicates assignment of the track $\mathbf{T}_i$ to a detection box $\mathbf{P}_j$.
Given the association score matrix from \eqref{eq:association_scores}, we compute the assignment matrix following \algref{algo:compute_associations} as described below.
In \gls{mot}, the number of tracks can differ from the number of detections. 
Therefore, we add a variable number of deletion elements $\mathbf{D}$ to $\mathbf{P}$, such that $\|\mathbf{P}\|  + \| \mathbf{D}\| \geq \|\mathbf{T}\|$ and each track in $\mathbf{T}$ can be assigned.

In this work, the parameters of the \gls{reid} model $\Phi$ are trained using our proposed self-supervised \gls{subco} task. 

 \begin{algorithm}[t]
\footnotesize
\SetAlgoLined
\KwData{\gls{reid} features $\mathbf{Y}^{t-1}$, $\mathbf{X}^{t}$, no match score threshold $\Delta_{match}$}
\KwResult{assignment matrix ${}^{t-1}_{}\mathbf{A}^{}_{t}$, deletion score vector ${}^{t-1}_{}\mathbf{d}^{}_{}$}
    ${}^{t-1}_{}\mathbf{S}^{}_{t} = \mathbf{Y}^{t-1} (\mathbf{X}^n)^T$\;
    
    \# Compute forward assignment scores, where: \\
    $\left[ \begin{array}{cc} {}^{t-1}_{}\mathbf{R}^{}_{t} & {}^{t-1}_{}\Vec{\mathbf{d}}  \end{array}\right] = 
    \psi_{j}\left( \left[ \begin{array}{cc} {}^{t-1}_{}\mathbf{S}^{}_{t} & \Vec{\mathds{1}} \cdot \Delta_{match} \\ \end{array}\right] \right)$ \;

    \# Compute backward assignment scores: \\
    $\left[ \begin{array}{c} {}^{t-1}_{}\mathbf{C}^{}_{t} \\ {}^{t-1}_{}\Vec{\mathbf{i}}  \end{array}\right] = 
    \psi_{i}\left( \left[ \begin{array}{c} {}^{t-1}_{}\mathbf{S}^{}_{t} \\ \Vec{\mathds{1}}^T \cdot \Delta_{match} \\ \end{array}\right] \right)$
    
    \# Compute assignment matrix by element-wise min of forward and backward assignment scores: \\
    \begin{equation}
        [{}^{t-1}_{}\mathbf{A}^{}_{t}]_{ij} = \min \left\{ [{}^{t-1}_{}\mathbf{R}^{}_{t}]_{ij}, [{}^{t-1}_{}\mathbf{C}^{}_{t}]_{ij}\right\}
        \label{eq:elementwise-minimum}
    \end{equation}
\caption{Assignment matrix computation.}
\label{algo:compute_associations}
\end{algorithm}

{\parskip=3pt
\noindent\textit{Assignment Matrix Computation}: We derive the assignment matrix $\mathbf{A} = \psi (\mathbf{S}) = \psi \left( \mathbf{Y} \left( \mathbf{X}\right)^T \right)$ from the association scores $\mathbf{S}$ and an assignment operation $\psi$.
During tracking inference, the non-differentiable Hungarian algorithm is chosen as $\psi$ to find a cost-optimal matching given the association scores. 
For end-to-end training, however, we choose $\psi$ as a differentiable function, which follows~\cite{liu2022utrack} and is designed to approximate a one-to-one matching.

First, we compute association scores between detection boxes of a frame pair following~\eqref{eq:association_scores}, followed by row-wise softmax normalization.
The forward assignment matrix is then computed by applying row-wise softmax activation to the association score matrix, as

{\small
\begin{align}
    [\mathbf{R}]_{i,j} = \psi_{i}(\mathbf{S}) = \frac{\tau\exp\left([\mathbf{S}]_{i,j}\right)}{\exp \left( \tau \Delta_{match}\right) + \sum_{j'} \exp \left( \tau[\mathbf{S}]_{i,j'}\right)}
\end{align} 
}

We accounted for starting and finished tracks that do not have correspondences in both frames by adding an absolute matching threshold $\Delta_{match}$ to the denominator. The association  score between the track and any detection feature embeddings has to exceed this threshold. Otherwise, the track is added to the set of deleted tracks. The row-wise softmax activation normalizes all association scores for the track features $\Vec{\mathbf{t}}_i$.
To prevent the model from assigning all detections $\mathbf{p}_j$ to the same track, we next compute backward assignment scores by applying column-wise softmax $\mathbf{C} = \psi_{j}(\mathbf{S})$. The final assignment matrix is then constructed from the elementwise minimum score from forward and backward assignment scores, following \eqref{eq:elementwise-minimum}. 

{\parskip=3pt
\noindent\textit{\gls{subco} Loss}: We propose formulating the self-supervised training signal as the consistency value between a sequence of propagated assignments and the assignments computed directly between the start and end frames. 
Hence, we interpret the assignment matrices as the adjacency matrix of the matching graph between two frames.
We initialize the tracks with the detections in the first frame $\mathbf{Y}^1 = \mathbf{X}^1$.
As we string the pairwise adjacency matrices of directed graphs along a sequence, the product of two matrices ${}^{t-1}_{}\mathbf{A}^{}_{t}$ and ${}^{t}_{}\mathbf{A}^{}_{t+1}$ describes how to get from vertices according to $\mathbf{P}^{t-1}$ to vertices from $\mathbf{P}^{t+1}$ following edges that go through $\mathbf{P}^{t}$.
By keeping the order of detection boxes in $\mathbf{P}^{t}$ unchanged for the computation of all association score matrices, we can trace tracks along these graphs up to the end of the sequence or their deletion by the matrix product of all adjacency matrices:
$  {}^{1}_{}\Tilde{\mathbf{A}}^{}_{T} = \prod_{t=2}^{T} {}^{t-1}_{}\mathbf{A}^{}_{t} $}.

The \gls{subco} loss function is then given by the negative log-sum of the row-wise scalar product between the propagated assignment matrix along short timescales ${}^{1}_{}\Tilde{\mathbf{A}}^{}_{T}$ and the direct assignment matrix between the start and end frame ${}^{1}_{}\mathbf{A}^{}_{T}$ for all non-deleted tracks, as follows:
\begin{align}
    \label{eq:inter-frame-loss}
    \mathcal{L}^{SubCo} = &\frac{1}{ \sum_i [{}^{1}_{}\mathbf{d}_i = 0 ] } \cdot \\ &\sum_{i} \begin{cases}
        0 & \text{if } {}^{1}_{}\mathbf{d}_i = 1 \\
                -\log \left( \sum_{j} [{}^{1}_{}\Tilde{\mathbf{A}}^{}_{T}]_{ij} \cdot [{}^{1}_{}\mathbf{A}^{}_{T}]_{ij} \right)  & \text{otherwise }, \\
    \end{cases} \nonumber
\end{align}
where $[\dots]$ denote the Iverson brackets. The overall loss computation pipeline is summarized in \algref{algo:compute_loss}.
\begin{algorithm}[t]
\footnotesize
\SetAlgoLined
\KwData{\gls{reid} features $\mathbf{X}^{1}, \dots, \mathbf{X}^{t}$, no match score threshold $\Delta_{match}$}
\KwResult{Inter frame loss value $\mathcal{L}^{inter}$}
    
    \# Compute association matrix over long timescale \\
    ${}^{1}_{}\mathbf{A}^{}_{T}, {}^{t-1}_{}\Tilde{\mathbf{d}} = \text{\algref{algo:compute_associations}} \left( \mathbf{X}^{1}, \mathbf{X}^{T}, \Delta_{match} \right)$ \;

    \# Iterate over short timescale associations: \\
    \For{$t \in [2, \dots, T]$}{
        \# Compute association matrix and deletion vector\\
        ${}^{t-1}_{}\mathbf{A}^{}_{t}, {}^{t-1}_{}\mathbf{d} = \text{\algref{algo:compute_associations}} \left( \mathbf{X}^{t-1}, \mathbf{X}^{t}, \Delta_{match} \right)$ \;
    }
    \# Propagate short time-scale association matrices \\
    ${}^{1}_{}\Tilde{\mathbf{A}}^{}_{T} = \prod_{t=[2, \dots, T]} {}^{t-1}_{}\mathbf{A}^{}_{t}$ \;

    \# Sum the deletion vectors associated with detections in the t=1 frame \\
    ${}^{1}_{}\mathbf{d} = \sum_{t=[2, \dots, T]}  \left( \prod_{t=[2, \dots, t]} {}^{t-1}_{}\mathbf{A}^{}_{i} \right)  {}^{t}_{}\mathbf{d}$ \;
    
    Compute the $\mathcal{L}^{inter}$ value by \equref{eq:inter-frame-loss}
    
\caption{Inter-frame loss computation.}
\label{algo:compute_loss}
\end{algorithm}

The motivation for enforcing consistency between association matrices over short and long-timescale association matrices is twofold: 
\begin{enumerate*}[label={\arabic*)}]
    \item We expect an object to be more visually similar between two frames with a smaller temporal distance since we assume that, for the most part, variations of scene parameters such as illumination, camera perspective, and position of objects increase with time. The \gls{reid} features should be invariant to such changes since they keep the object's identity unaffected. Therefore, we add a long-timescale comparison with assignment scores between the start and end frames.
    \item The \gls{reid} features should further be discriminative between object identities. Therefore, due to the multistep propagation, the associations are rewarded for having high scores for a unique path through the sequence due to the multiplicative propagation.
\end{enumerate*}

By excluding deleted tracks from the loss computation, we give way to the degenerate case, where the \gls{reid} model converges to delete all tracks.  
Therefore, we enforce keeping at least one track alive. This is justified by the characteristics of the data, that the intersection of instance IDs between all frames is strictly nonzero in all annotated examples of the dataset, which we study more in-depth in the supplementary material.\looseness=-1

{\parskip=3pt
\noindent\textit{Intra-Frame Loss}: We complement our loss function with an intra-frame loss term that encourages maximizing the dissimilarity of \gls{reid} features within the same frame:
\begin{align}
    \mathcal{L}^{intra} =  \sum_{t = 1 \dots T} \frac{1}{(K^t)^2} \| {}^{t}_{}\mathbf{A}^{}_{t} - \mathbf{I} \|_1
\end{align}
This loss term is motivated by the problem formulation, which expects all objects in the same frame to have unique identities.

\subsection{Model Architecture} 

{\parskip=3pt
\noindent\textit{Object detector}: We use YOLO-X~\cite{yolox2021} as an object detector in its large configuration. It is an anchor-free detector and extends the YOLO architecture by a multi-scale approach and decoupled classification and regression heads to improve accuracy and reduce the number of false positive detections. YOLO-X predicts a fixed set of default bounding boxes and their corresponding class probabilities. During detection, final boxes are selected based on non-maximum suppression and a confidence threshold.}

{\parskip=3pt
\noindent\textit{\gls{reid} model}: We experiment with a ResNet-50~\cite{he2016resnet} and a \gls{vit} architecture for the \gls{reid} model. ResNet-50 is the standard model used in supervised and self-supervised \gls{mot}, and \gls{vit} allows comparison to single-frame self-supervised pre-training strategies.}

{\parskip=3pt
\noindent\textit{Tracking logic}: 
\label{sec:tracking_logic}
During inference, we employ the BYTE tracklet association method ~\cite{zhang2022bytetrack}, which first associates high-scoring detection boxes $\mathcal{D}_{high}$ with open tracks $\tau$ by maximizing \gls{iou} between bounding boxes in $\mathcal{D}_{high}$ as well as the propagated bounding boxes of open tracks propagated by a Kalman filter. We combine these IoU similarities with our learned \gls{reid} features, similarly to FairMOT~\cite{zhang2021fairmot} as

\begin{align}
    c_{i,j} = IoU\left( \mathbf{b}_i, \Tilde{\mathbf{b}}_j \right) + \omega_{ReID} \frac{\mathbf{x}_i^T \mathbf{x}_j}{\|\mathbf{x}_i\| \cdot \| \mathbf{x}_j \|},
\end{align}
where $\Tilde{\mathbf{b}}_j$ is the predicted bounding box location for tracklet $j$ for the current frame by the Kalman filter.
A second association step then assigns the low detection boxes $\mathcal{D}_{low}$ to the remaining tracklets $\tau_{remain}$. The authors in \cite{zhang2022bytetrack} propose to use IoU costs alone for the second association step, motivated by the observation that \say{the low score detection boxes usually contain severe occlusion or motion blur and appearance features are not reliable.} However, we utilize the \gls{reid} features in both association steps for our experiments.

\section{Experimental Evaluation}
In this section, we evaluate \gls{reid} models trained using our proposed \gls{subco} method on the \gls{bdd} as well as MOT17 datasets by comparing their performance with other supervised and self-supervised pretrained trackers.
We perform all the experiments using the \textit{MMTracking} framework.

\subsection{Implementation Details}

{\parskip=3pt
\noindent\textit{Datasets}: We train and evaluate our methods on the \acrlong{mot17} and \acrfull{bdd} datasets. The MOT17 challenge consists of 14 video sequences (7 training, 7 test) at varying frame rates ($>$14 fps) and sequence lengths ($>$\SI{20}{\second}) in unconstrained environments filmed with both static and moving cameras. Only pedestrian tracks are annotated and evaluated.}

The \gls{bdd} dataset comprises 100,000 crowd-sourced videos of driving scenes in urban, rural, and highway environments. It contains recordings of different weather conditions and times of the day. 
We chose this dataset to evaluate the multi-class tracking performance of our method.
For \gls{mot} benchmarking, the \gls{bdd} dataset provides track identity annotations for eight traffic participant classes at 5 fps over 200 validation and 400 test videos of \SI{40}{\second} length. 

\begin{table*}[ht]
\scriptsize
\centering
\caption{Evaluation of supervised and self-supervised MOT methods on the BDD100k val set.  The respective references indicate external results. The remaining methods are evaluated using identical \textit{YOLO-X} detections for a fair comparison.}
\label{tab:bdd100k-val-results}
\begin{tabular}{llllllllllllll} \toprule
& Method & Detector & ReID & mHOTA & mMOTA & mIDF1 & mDetA & mAssA & MT & ML & IDSw \\ \midrule
 \multirow{3}{*}{\rotatebox[origin=c]{90}{MOT}} & ByteTrack & YOLOX-L & - & 42.5 & \underline{42.4} & 43.9 & \underline{41.0} & 46.5 & 6598 & 4605 & 47883 \\
& QDTrack & Faster R-CNN & RoI head & 41.7 & 35.6 & \underline{49.7} & 36.3 & \underline{51.5} & \textbf{8827} & \textbf{3175} & \textbf{6262} \\
& TETer~\cite{trackeverything} & Faster R-CNN & RoI head & - & - & \textbf{52.9} & 39.1 & \textbf{53.3} & - & - & - \\ \midrule
 \multirow{6}{*}{\rotatebox[origin=c]{90}{SSMOT}} & MoCov2 & YOLOX-L & R50 & 29.5 & 43.2 & 19.9 & 23.9 & 27.5 & 7538 & 3820 & 117264 \\
& SwAV & YOLOX-L & R50 & 28.5 & 43.9 & 19.0 & 22.9 & 26.5 & 7692 & 3791 & 117909 \\
& MaskFeat & YOLOX-L & ViT-B & 37.7 & 32.1 & 41.5 & 37.2 & 39.2 & 6646 & 4600 & 45534 \\
& MoCo v3 & YOLOX-L & ViT-S & 36.6 & 37.8 & 37.1 & 35.3 & 48.0 & 7842 & 3611 & 56658 \\
& SubCo-ResNet (Ours) & YOLOX-L & R50 & \underline{43.3} & \textbf{44.0} & 48.3 & \textbf{42.3} & 48.3 & \underline{7857} & \underline{3546} & \underline{37461} \\
& SubCo-ViT (Ours) & YOLOX-L & ViT-S & \textbf{43.6} & 42.3 & 46.1 & 39.2 & 49.5 & 7748 & 3634 & 40031 \\
 \bottomrule
\end{tabular}
\vspace{-0.2cm}
\end{table*}

\begin{table}[t]
\scriptsize
\centering
\caption{Evaluation of supervised and self-supervised tracking methods on the MOT17 \textit{half-val} benchmark. The respective references indicates external results, the remaining methods are evaluated using the same detection proposals obtained from a pre-trained \textit{YOLO-X} detector~\cite{zhang2022bytetrack} for a fair comparison.}
\label{tab:mot17-val-benchmark}
\begin{tabular}{lllllll}
\toprule
 & Method & ReID & MOTA & IDF1 & IDSw & HOTA \\
 \midrule
\multirow{5}{*}{\rotatebox[origin=c]{90}{MOT}} & SORT & - & 62.0 & 57.8 & 1947 & - \\
 & QDTrack & RoI & 59.1 & 71.6 & 867 & 59.1 \\
 & DeepSORT & R50 & 63.8 & 69.6 & 1061 & - \\
 & Tracktor & R50 & 64.7 & 66.6 & 1152 & - \\
 & ByteTrack & - & \textbf{78.5} & \textbf{78.3} & \textbf{219} & \textbf{67.1} \\ \midrule
  \multirow{10}{*}{\rotatebox[origin=c]{90}{SSMOT}} & FairMOT~\cite{liu2022utrack} & R50 & 65.8 & 61.0 & 1098 & - \\
 & SimpleReID~\cite{karthik2020simple} & R50 & 61.7 & 65.2 & - & - \\
 & SSL-MOT~\cite{kim2022sslmot} & R50 & 61.5 & 55.0 & 2957 & - \\
 & CenterTrack~\cite{kim2022sslmot} & - & 61.5 & 59.6 & 2583 & - \\
 & UTrack~\cite{liu2022utrack} & R50 & 67.6 & 71.8 & 503 & - \\
 & MoCov2-BYTE & R50  & 62.6 & 66.1 & 654 & 58.9 \\
 & MaskFeat-BYTE & ViT-B & 47.5 & 54.7 & 757 & 49.3 \\
 & MoCo v3-BYTE & ViT-S  & 62.3  & 66.1 & 785 & 56.8\\
 & SubCo-BYTE (Ours) & R50 & \underline{77.0} & \underline{77.5} & \underline{493} & \underline{66.3} \\
 & SubCo-BYTE (Ours) & ViT-S & 76.5 & 76.7 & 571 & 61.3 \\ 
 \bottomrule
\end{tabular}
\vspace{-0.5cm}
\end{table}

{\parskip=3pt
\noindent\textit{Proposal generation}: We use the YOLO-X~\cite{yolox2021} object detector to generate object proposals for training and evaluation of the \gls{reid} models. We chose YOLO-X as the base detector to directly compare against ByteTrack, as a strong parameter-free tracker on both benchmarks.
For \gls{bdd}, we train YOLO-X for 20 epochs using 4 NVIDIA V100 GPUs on the \gls{mot} and detection annotations. All other training hyperparameters are identical to \textit{ByteTrack}~\cite{zhang2022bytetrack}. For the MOT17 evaluation, we rely on the pre-trained detector provided by \textit{ByteTrack}~\cite{zhang2022bytetrack}.}

{\parskip=3pt
\noindent\textit{\gls{reid} model and training}: All the \gls{reid} models are trained on 4 GPUs. The batch size varies with the sequence length for the proposed loss function. We train our methods on predictions from the YOLO-X~\cite{yolox2021} detector with confidence scores of $\geq 0.2$. Thus, we include artifacts such as false positive (high-scoring detection boxes but no instances) and false negative (object instances not detected) detections already in the training phase. For training, we use all the \gls{bdd} videos from which we extract frames at 5~Hz, excluding the validation and test set videos. For MOT17, we follow the common practice of splitting each training sequence into two halves and using the first half frames for training and the second for validation~\cite{zhou2020centertrack}. We train the \gls{reid} models for 20 epochs using the AdamW optimizer. The initial learning rate is set $2 \cdot 10^{-4}$ and is reduced by a factor of 10 at the 12th epoch.}

\subsection{Benchmark Evaluation}

We use the CLEAR~\cite{bernardin2008clear} and HOTA~\cite{luiten2020hota} metrics for evaluation to highlight different aspects of the tracking performance. 
\Gls{mota} combines FP, FN, and IDSw into a single scalar, while IDF1 quantifies the identity preservation ability of a tracker.
\Gls{hota} is a recently proposed metric that explicitly balances the effect of performing accurate detection (DetA), association (AssA), and localization (LocA)~\cite{luiten2020hota}. For the \gls{bdd} dataset, we report multi-class metrics, such as mMOTA and mHOTA, that are computed by averaging the MOTA / HOTA of each class type.


\subsubsection{BDD100k MOT Results}

We compare the \gls{mot} tracking performances of various trackers on the \gls{bdd} \textit{val} benchmark in \tabref{tab:bdd100k-val-results}.
The baselines include the \gls{sota} supervised trackers for the \gls{bdd} benchmark.
Additionally, we provide results for other self-supervised methods, that have identical architecture as our \gls{subco} models but vary in the training strategy for the \gls{reid} model, \ie MoCov2~\cite{chen2020improved}, SwaV~\cite{caron2020swav}, Mocov3~\cite{chen2021mocov3}, and MaskFeat~\cite{wei2022maskedfeat}.

The BYTE tracker using a ResNet-50 as \gls{reid} model with weights trained on the \gls{subco} task achieves the highest tracking accuracy of $42.3\%$ for \gls{mota} averaged over all class types. It thereby outperforms all supervised and unsupervised baselines by $>1.1\%$ in the mMOTA score. It also achieves the second-highest mHOTA score of $43.3\%$.
The highest mHOTA score of $43.6\%$ the tracker configuration using a \gls{vit} network as \gls{reid} model, with weights trained annotation-free on the \gls{subco} task.

\subsubsection{MOT17 Results}

\begin{table*}[ht]
\centering
\scriptsize
\caption{Ablation study of varying hyperparameter settings of the proposed loss function.}
\label{tab:ablation_loss_config}
\begin{tabular}{ccc|ccccc|ccccc} \toprule
 &  &  & \multicolumn{5}{c|}{MOT 17 half-val} & \multicolumn{5}{c}{BDD100k val} \\
 \cmidrule{4-13}
Seq length & Inter frame & Intra frame & HOTA & AssA & IDF1 & MOTA & IDSw & mHOTA & mAssA & mIDF1 & mMOTA & IDSw \\ \midrule
T=1 &  & \checkmark & 13.4 & 3.3 & 4.1 & -1.1 & 7715 & 13.5 & 3.3 & 20.0 & -4.3 & 143930 \\
T=4 & \checkmark &  & 63.2 & 61.2 & 73.9 & 76.3 & 587 & 41.3 & 41.7 & 44.2 & 34.2 & 56658 \\
T=4 & \checkmark & \checkmark & 65.0 & 64.3 & 75.5 & 76.8 & 481 & 42.3 & 43.3 & 43.0 & 36.7 & 47501 \\
T=8 & \checkmark &  & 64.2 & 62.6 & 74.8 & 76.6 & 522 & \textbf{44.0} & \textbf{48.9} & 46.4 & 37.5 & 41929 \\
T=8 & \checkmark & \checkmark & \textbf{66.3} & \textbf{66.4} & \textbf{77.5} & \textbf{77.0} & \textbf{219} & 43.3 & 48.3 & \textbf{48.3} & \textbf{42.3} & \textbf{37461} \\ \bottomrule
\end{tabular}
\end{table*}

In \tabref{tab:mot17-val-benchmark}, we compare results for the MOT17 \textit{half-val} benchmark to both supervised (\textit{MOT}) and self-supervised (\textit{SSMOT}) baselines from the literature.
Interestingly, ByteTrack which uses only \gls{iou}-based cost metrics for association matching, outperforms all the \gls{reid} methods by $+1.5\%$ in terms of \gls{mota} and $+0.8\%$ in terms of IDF1.
These results align with the observations by \textit{Zhang et al.}~\cite{zhang2022bytetrack} for the MOT17 dataset.
We assume that due to the high frame rate of the dataset, the location and constant velocity assumptions made by the \gls{iou} based association methods are accurate, while \gls{reid} performance suffers from many small and, thus, low-resolution detections.
The \gls{reid} model pre-trained using our proposed method outperforms all supervised and unsupervised \gls{reid} models in terms of ID F1 score and achieves the lowest number of identity switches, which indicates strong consistency of the \gls{reid} features along the entire trajectory.

\subsection{Ablation Study}

\tabref{tab:ablation_loss_config} summarizes the ablation study over sequence length and loss terms to give insights into how various hyperparameters in our loss formulation interact and contribute to the final benchmark performance. Please note that the negative \gls{mota} value in the first row results from a high number of false positives paired with many identity switches, such that the number of errors exceeds the number of detections in the ground truth. These metrics imply that the intra-frame loss cannot learn meaningful \gls{reid} features.
Combined with our proposed inter-frame loss, it positively affects consistent feature representation and boosts by $+1.6\%$ IDF1 for a sequence length of 4 and by $+2.7\%$ IDF1 for a sequence length of 8 during training.
Adding the inter-frame loss even outperforms training with twice the sequence length on the inter-frame loss.
This outcome is remarkable, as the sequence length of one data sample has to fit on a single GPU during the computation of the proposed loss function.
The intra-frame loss relaxes the trade-off of sequence length and model performance.

The ablation experiments presented in \tabref{tab:ablation_loss_config} further indicate that longer sequence lengths during training boost tracking performance of the learned \gls{reid} features. 
This result manifests mainly in the reduced number of ID switches and increased association accuracy of more than $+2\%$ when doubling the sequence length from 4 to 8.

{\parskip=3pt
\noindent\textit{Tracking logic experiments}: In \tabref{tab:ablation_tracking_features}, we compare the learned \gls{reid} features against \gls{iou}-based association matching in varying configurations of the BYTE~\cite{zhang2022bytetrack} tracking logic.
Therefore, we perform the first matching with high-scoring detection boxes (\#1) and the second matching of low-scoring detection boxes with tracking boxes by either maximizing \glspl{iou} or \gls{reid} feature similarities. 
A low detection score might result from poor visibility of an object, such as partial occlusion, small object sizes, or motion blur. We are especially interested in motion-based (IoU-based association) performance compared to our feature-based approach under these circumstances.
We found from the benchmarking experiments on the MOT17 dataset presented in \tabref{tab:mot17-val-benchmark}, that pure \gls{iou}-based matching outperforms pure \gls{reid}-based matching by $+1.5\%$ in the \gls{mota} score, due to the high frame rate in the dataset, where the constant velocity assumption mostly holds. However, the \gls{iou}-based tracking generates three times more ID switches, presumably caused by similar occlusion scenarios to the ones shown in \figref{fig:qualitative-example-mot}.}

\begin{table}[ht]
\scriptsize
\centering
\caption{Comparison of different combinations of components for association costs used in the first and second association stage of BYTE~\cite{zhang2022bytetrack} on the MOT17 and BDD100k validation sets.}
\label{tab:ablation_tracking_features}
\begin{tabular}{ll|ccc|ccc} \toprule
 &      & \multicolumn{3}{c|}{MOT17 half-val} & \multicolumn{3}{c}{BDD100k val} \\
 \cmidrule{3-8}
\#1  & \#2  & MOTA    & IDF1    & IDSw   & mMOTA    & mIDF1   & IDSw   \\
\midrule
IoU & ReID & 75.0 & 76.3 & 528 & 37.5 & \textbf{48.9} & 41929 \\
IoU & IoU & \textbf{78.5} & 76.5 & 692 & 36.1 & 46.5 & 47883 \\
ReID & ReID & 77.0 & \textbf{78.3} & \textbf{219} & \textbf{42.3} & 48.3 & \textbf{37461} \\
ReID & IoU & 76.3 & 73.9 & 587 & 32.8 & 42.8 & 41929 \\ \bottomrule
\end{tabular}
\end{table}
\begin{table}[t]
\scriptsize
\centering
\caption{Comparison of different data association methods on MOT17 and BDD100K val set using \gls{reid} features learned using our \gls{subco} method.}
\label{tab:ablation_trackers}
\begin{tabular}{ll|ll|ll}
\toprule
& & \multicolumn{2}{c|}{MOT17  half-val} & \multicolumn{2}{c}{BDD100k val} \\
 \cmidrule{3-6}
Detections & Tracker logic & MOTA & \multicolumn{1}{l|}{IDF1} & mMOTA & mIDF1 \\  
\midrule
YOLOX & DeepSORT & 63.8 & 69.6 & 30.2 & 39.9 \\
CenterNet~\cite{zhang2021fairmot} & FairMOT~\cite{zhang2021fairmot} & 65.8 & 61.0  & - & - \\
\midrule
YOLOX & DeepSORT+\gls{subco} & 64.6 & 74.4  & 36.6 & 44.0 \\
YOLOX & BYTE+\gls{subco} & \textbf{77.0} & \textbf{78.3} & \textbf{42.3} & \textbf{48.3} \\ 
CenterNet~\cite{zhang2021fairmot} & BYTE+\gls{subco} & 74.6 & 77.8  & - & - \\
\bottomrule
\end{tabular}
\vspace{-0.4cm}
\end{table}

\begin{figure*}[ht]
    \centering
    \begin{subfigure}[b]{0.24\textwidth}
        \centering
        \includegraphics[trim=0 20 0 20,clip,width=\textwidth]{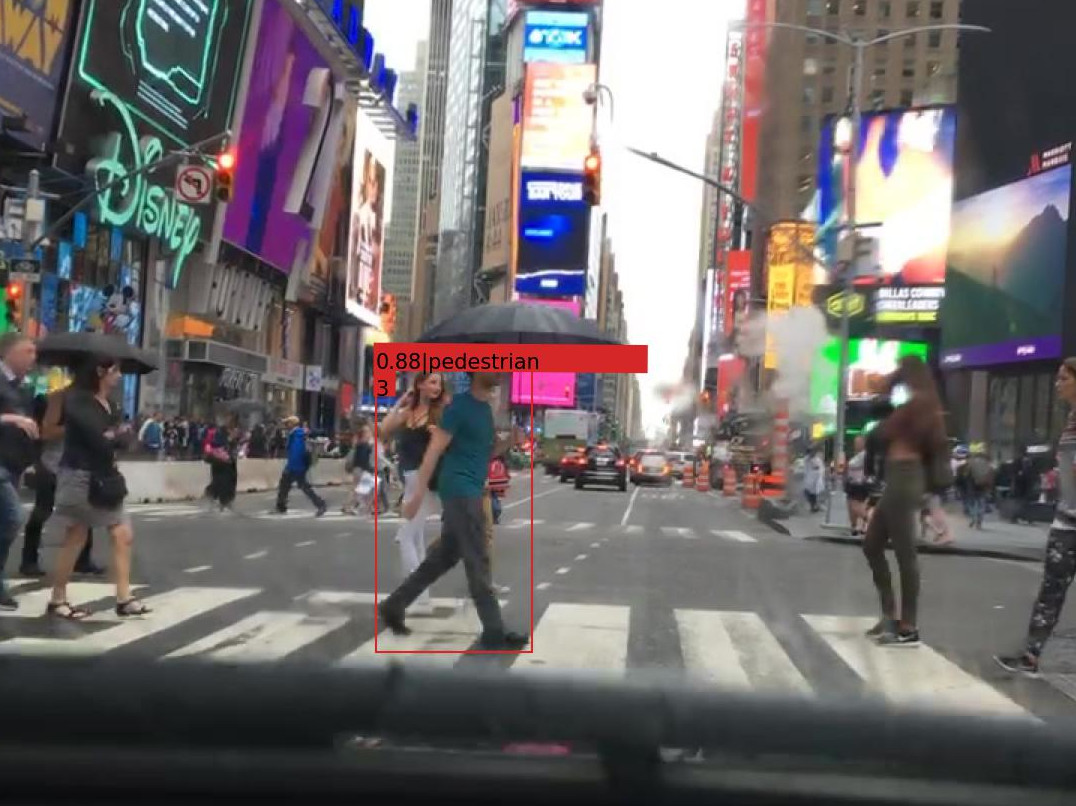}
    \end{subfigure}
    \hfill
    \begin{subfigure}[b]{0.24\textwidth}
        \centering
        \includegraphics[trim=0 20 0 20,clip,width=\textwidth]{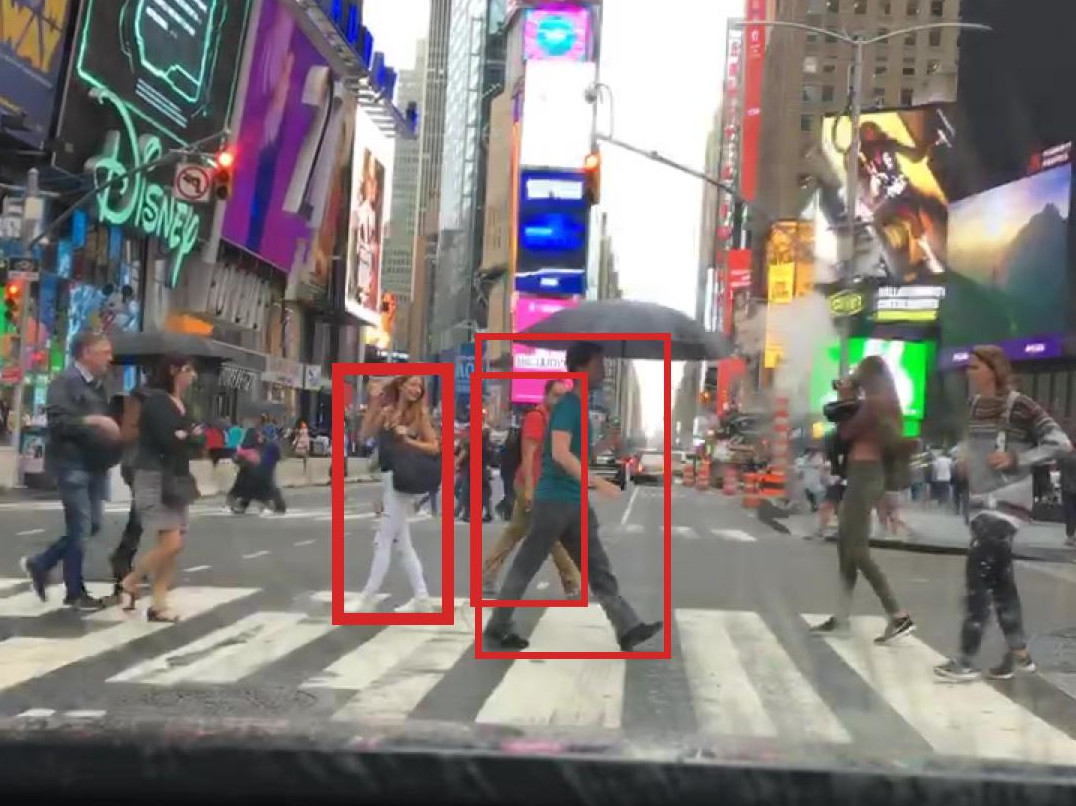}
    \end{subfigure}
    \hfill
    \begin{subfigure}[b]{0.24\textwidth}
        \centering
        \includegraphics[trim=0 20 0 20,clip,width=\textwidth]{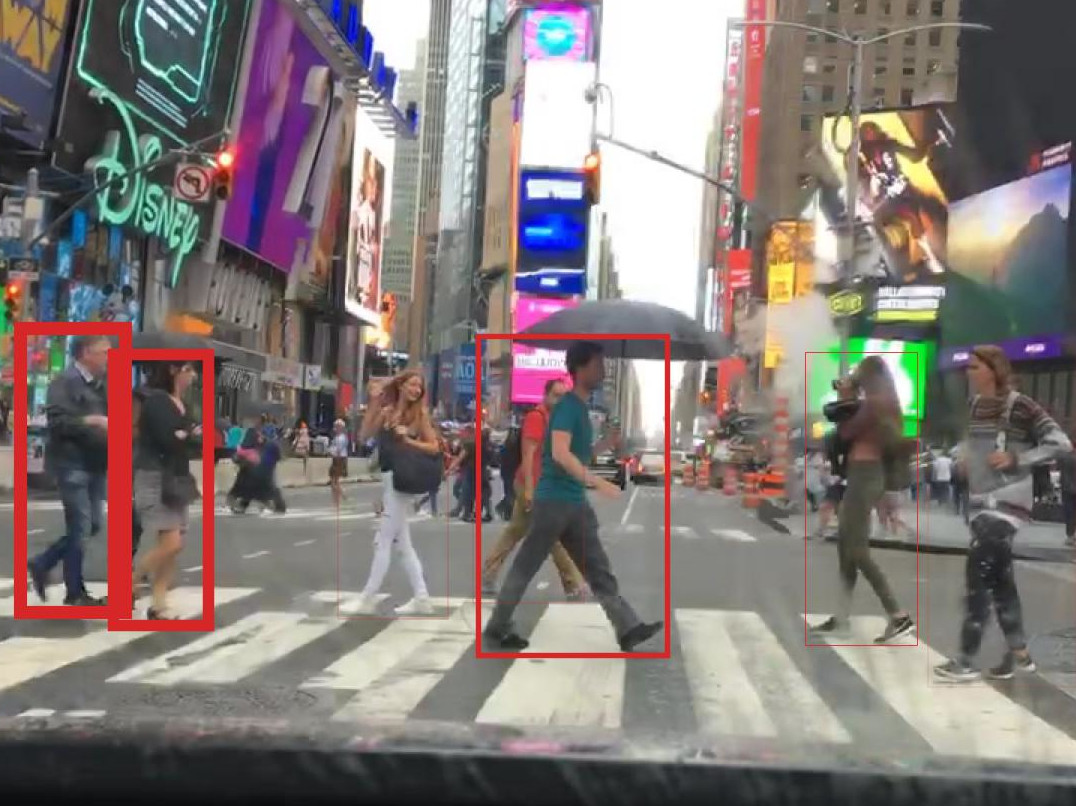}
    \end{subfigure}
    \hfill
    \begin{subfigure}[b]{0.24\textwidth}
        \centering
        \includegraphics[trim=0 20 0 20,clip,width=\textwidth]{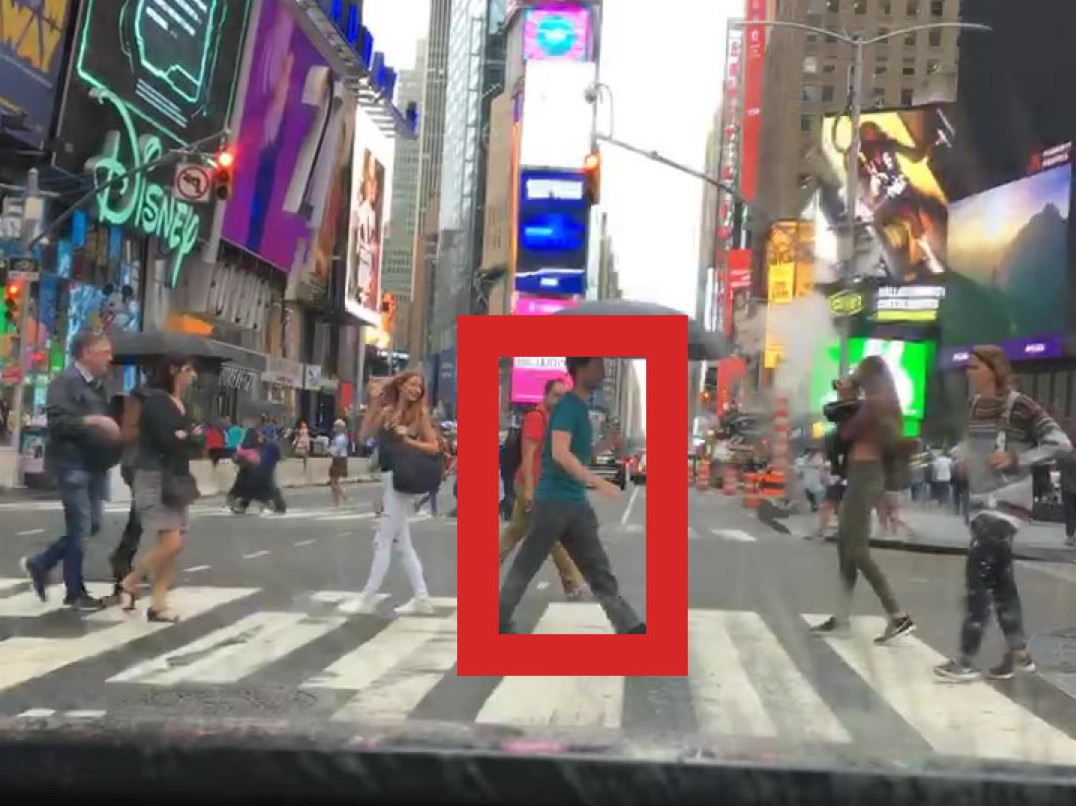}
    \end{subfigure}
    
    \vspace{0.2cm}
    
    \begin{subfigure}[b]{0.24\textwidth}
        \centering
        \includegraphics[trim=0 20 0 20,clip,width=\textwidth]{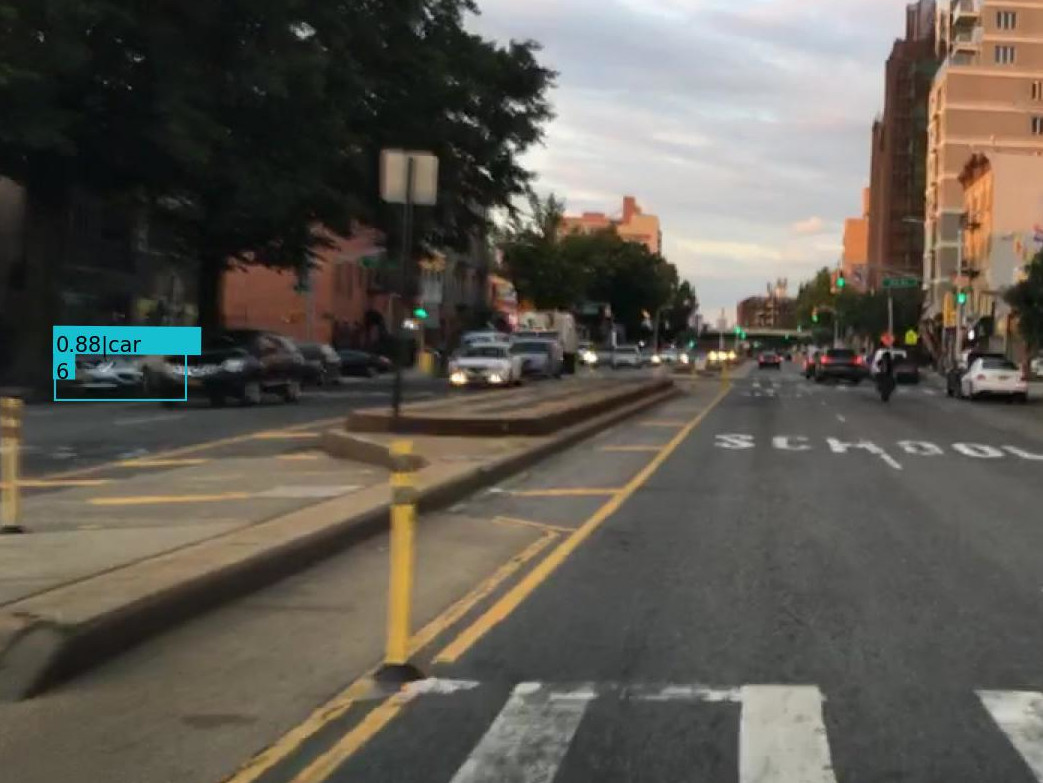}
    \end{subfigure}
    \hfill
    \begin{subfigure}[b]{0.24\textwidth}
        \centering
        \includegraphics[trim=0 20 0 20,clip,width=\textwidth]{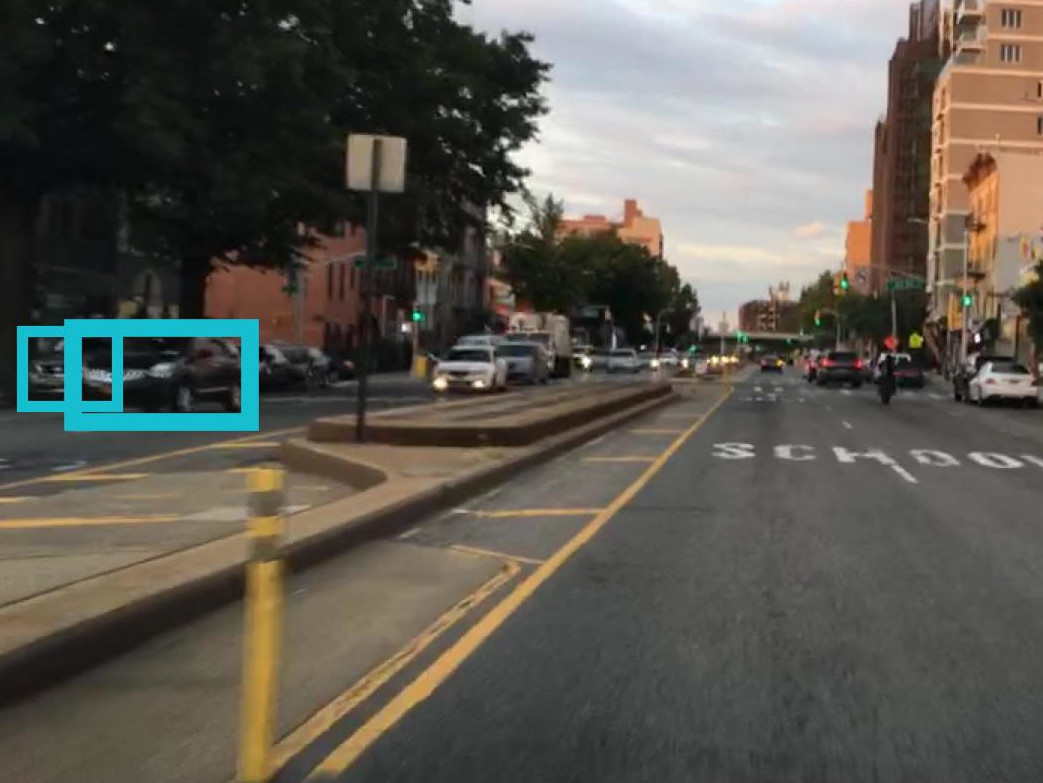}
    \end{subfigure}
    \hfill
    \begin{subfigure}[b]{0.24\textwidth}
        \centering
        \includegraphics[trim=0 20 0 20,clip,width=\textwidth]{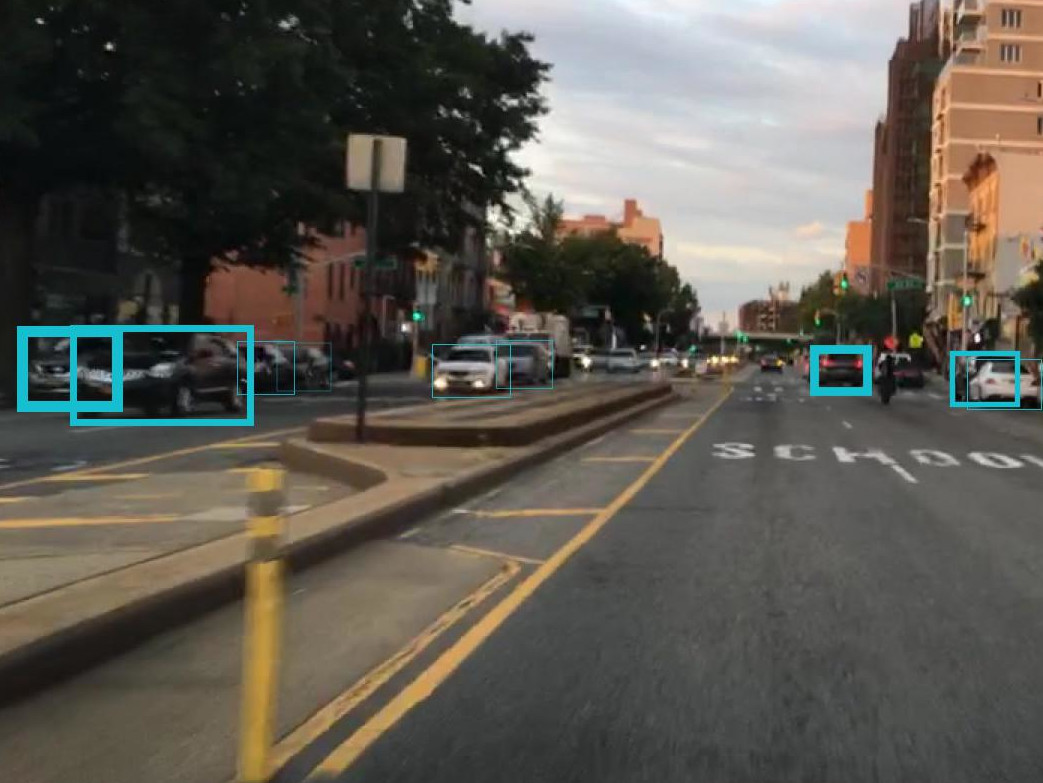}
    \end{subfigure}
    \hfill
    \begin{subfigure}[b]{0.24\textwidth}
        \centering
        \includegraphics[trim=0 20 0 20,clip,width=\textwidth]{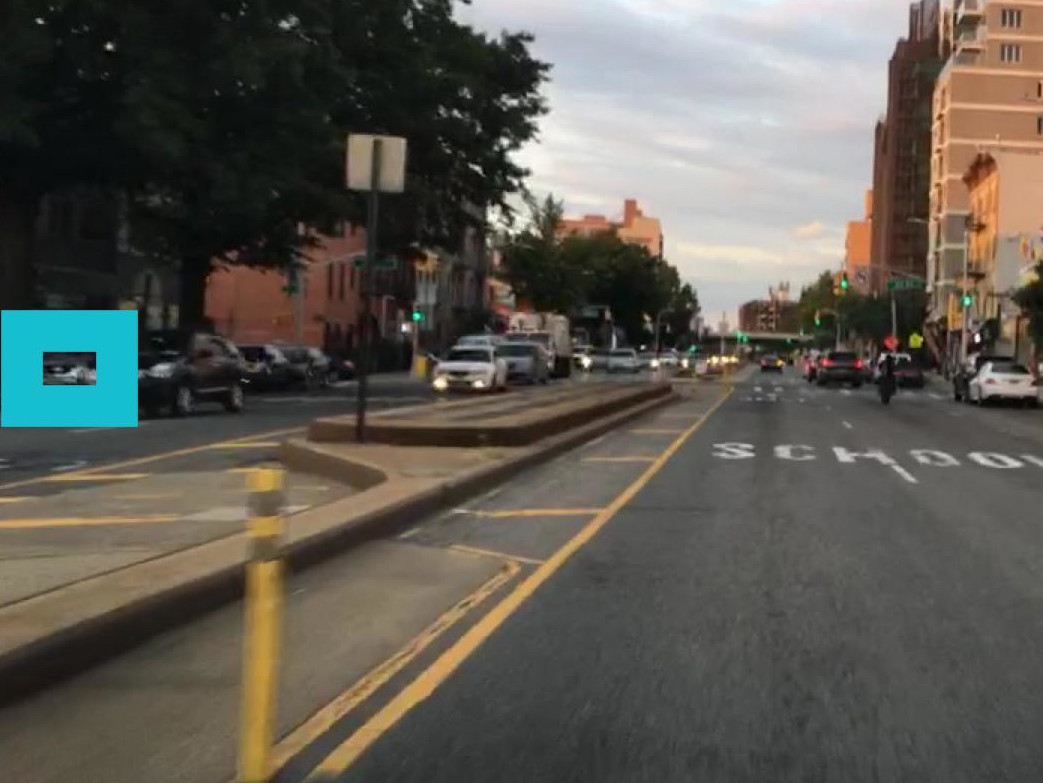}
    \end{subfigure}

    \vspace{0.2cm}

    \begin{subfigure}[b]{0.24\textwidth}
        \centering
        \includegraphics[trim=0 20 0 20,clip,width=\textwidth]{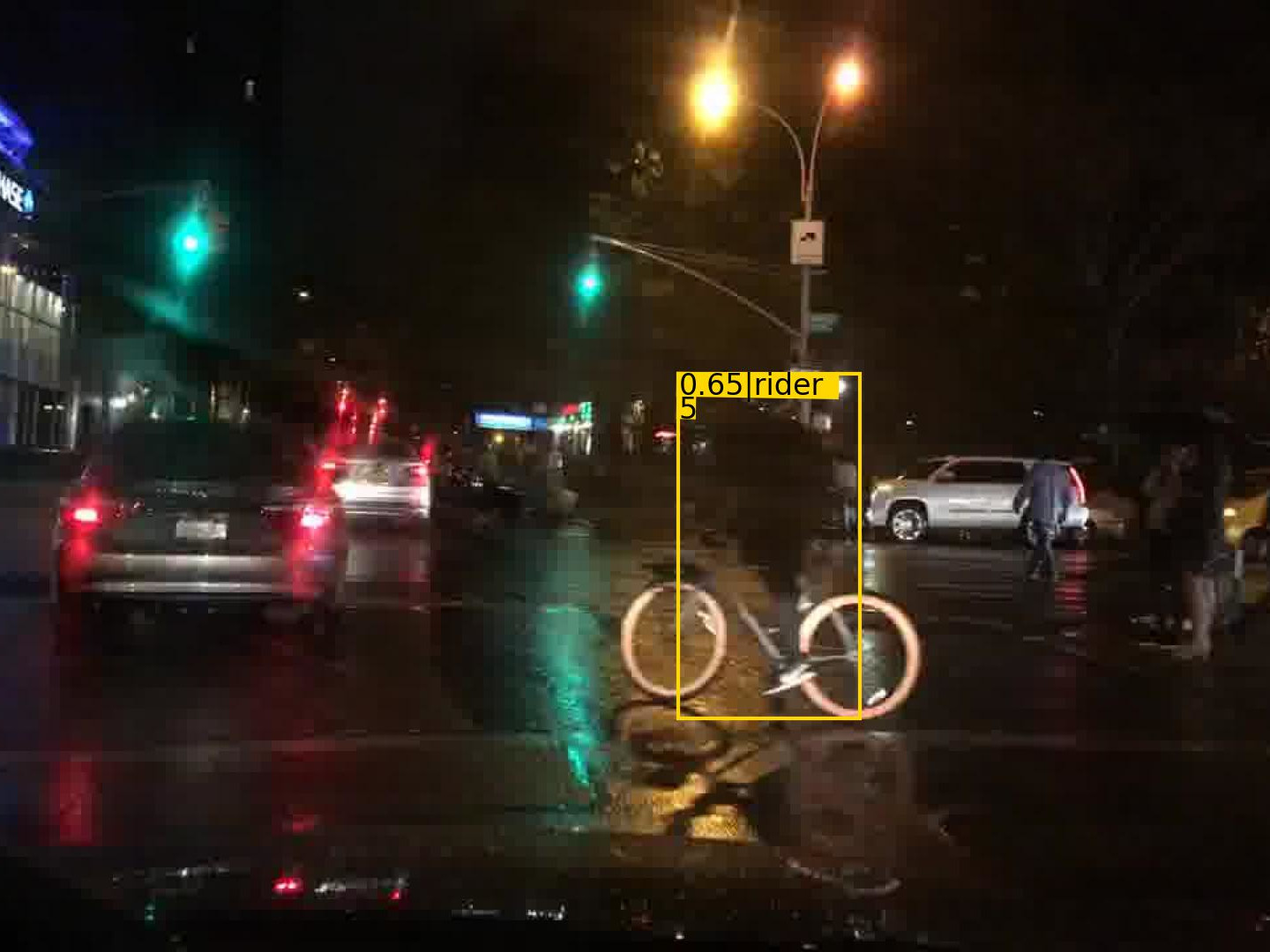}
        \caption{Reference frame}
    \end{subfigure}
    \hfill
    \begin{subfigure}[b]{0.24\textwidth}
        \centering
        \includegraphics[trim=0 20 0 20,clip,width=\textwidth]{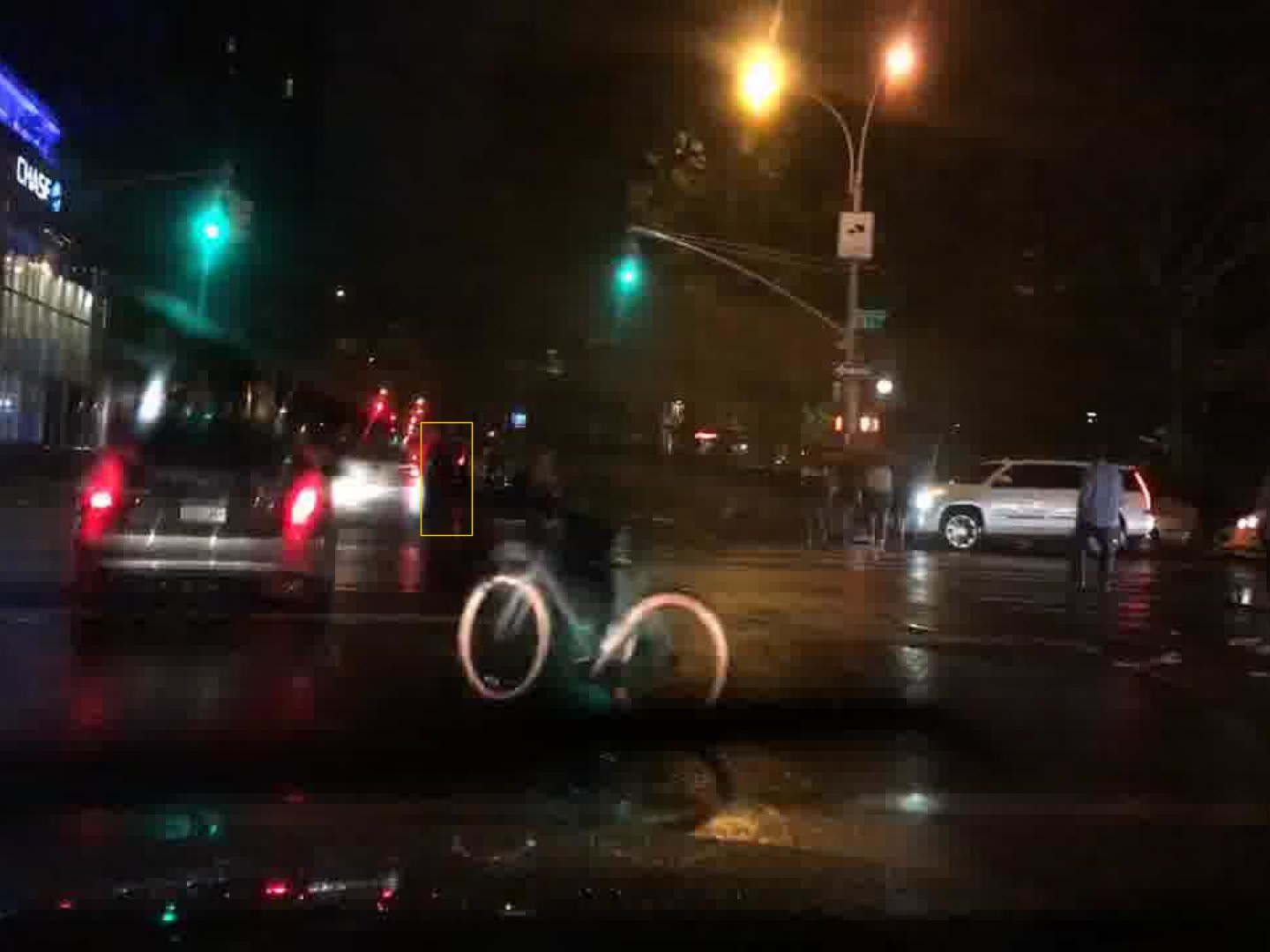}
        \caption{Bbox IoU}
    \end{subfigure}
    \hfill
    \begin{subfigure}[b]{0.24\textwidth}
        \centering
        \includegraphics[trim=0 20 0 20,clip,width=\textwidth]{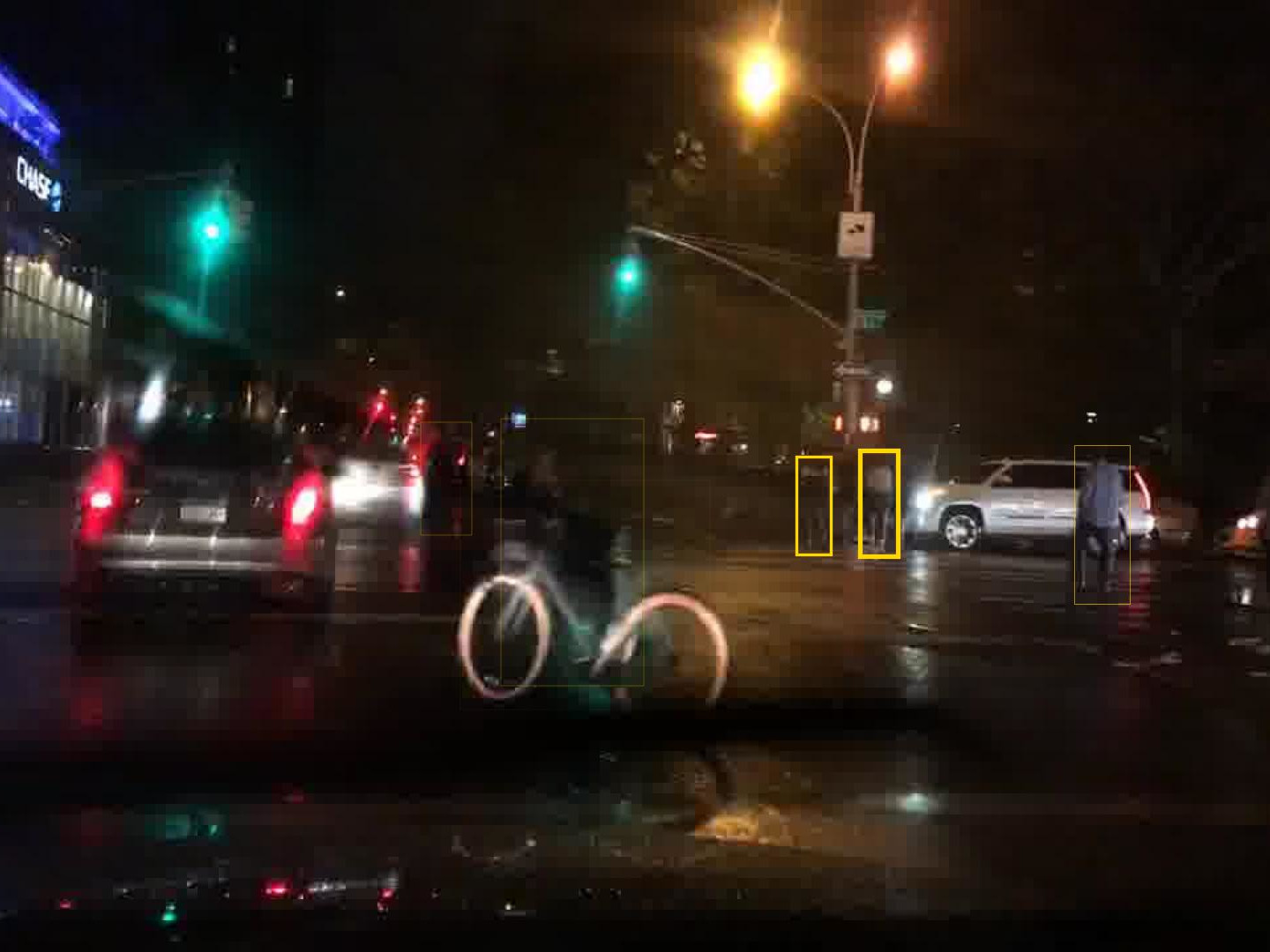}
        \caption{SwaV ReID}
    \end{subfigure}
    \hfill
    \begin{subfigure}[b]{0.24\textwidth}
        \centering
        \includegraphics[trim=0 20 0 20,clip,width=\textwidth]{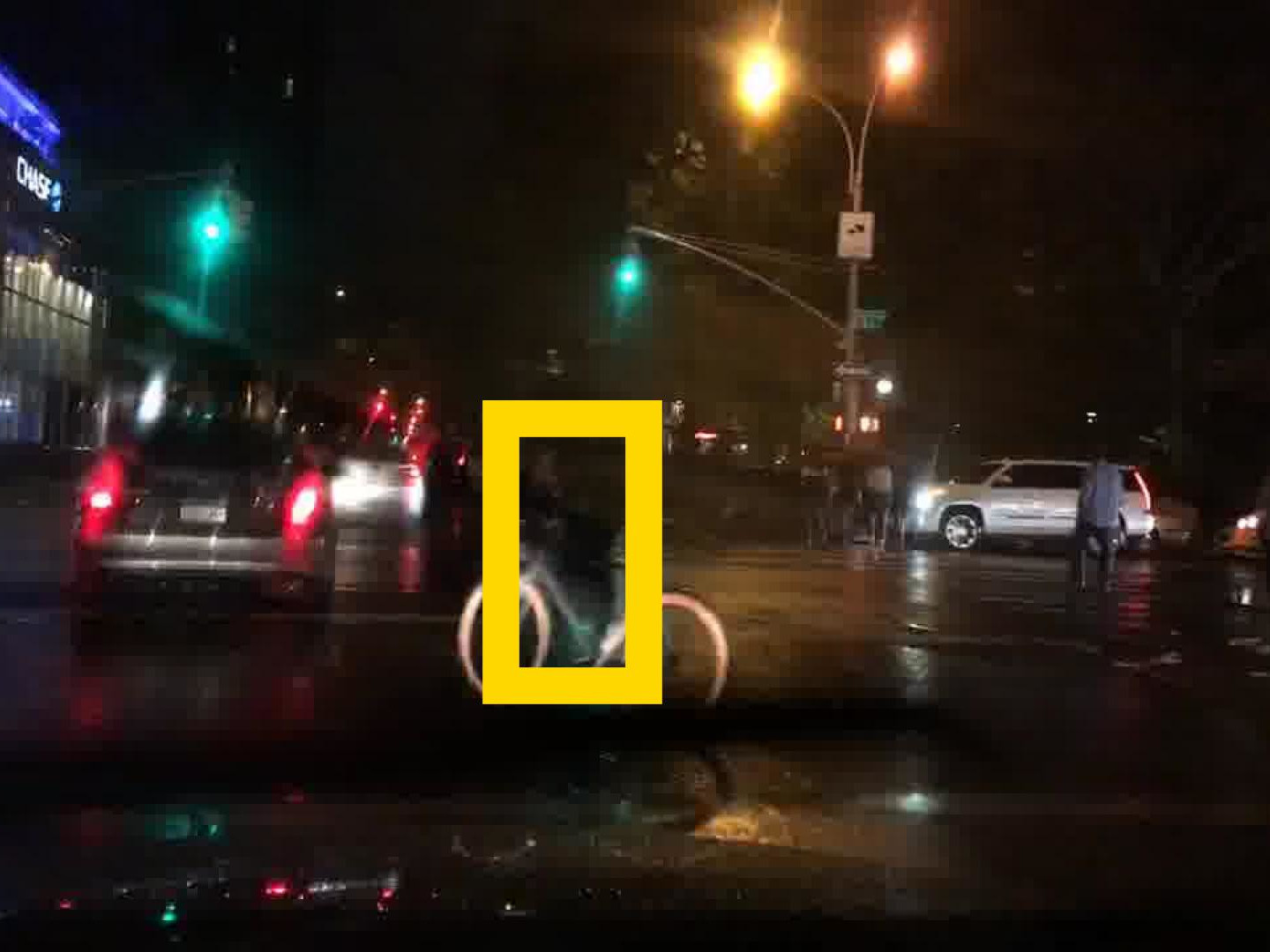}
        \caption{SubCo ReID (Ours)}
    \end{subfigure}

    \caption{Qualitative comparison of association methods for selected sequences on the BDD100k dataset. The figures in (a) display one detection box in the reference frame, color-coded by the assigned object identity. Figures (b)-(d) show detection boxes in the subsequent frame, where the line thickness indicate the association strengths to detection boxes in the reference frame by different re-identification approaches. These comprise (b) IoU of the propagated reference boxes using a Kalman filter as done in ByteTrack~\cite{zhang2022bytetrack}, as well as dot-product feature similarity between (c) SwaV extracted features and (d) ReID features extracted using our proposed approach.}
    \label{fig:qualitative-example-mot}
    \vspace{-8pt}
\end{figure*}

\begin{figure*}[ht]
    \centering
    \begin{subfigure}[b]{0.24\textwidth}
        \centering
        \includegraphics[trim=0 0 0 0,clip,width=\textwidth]{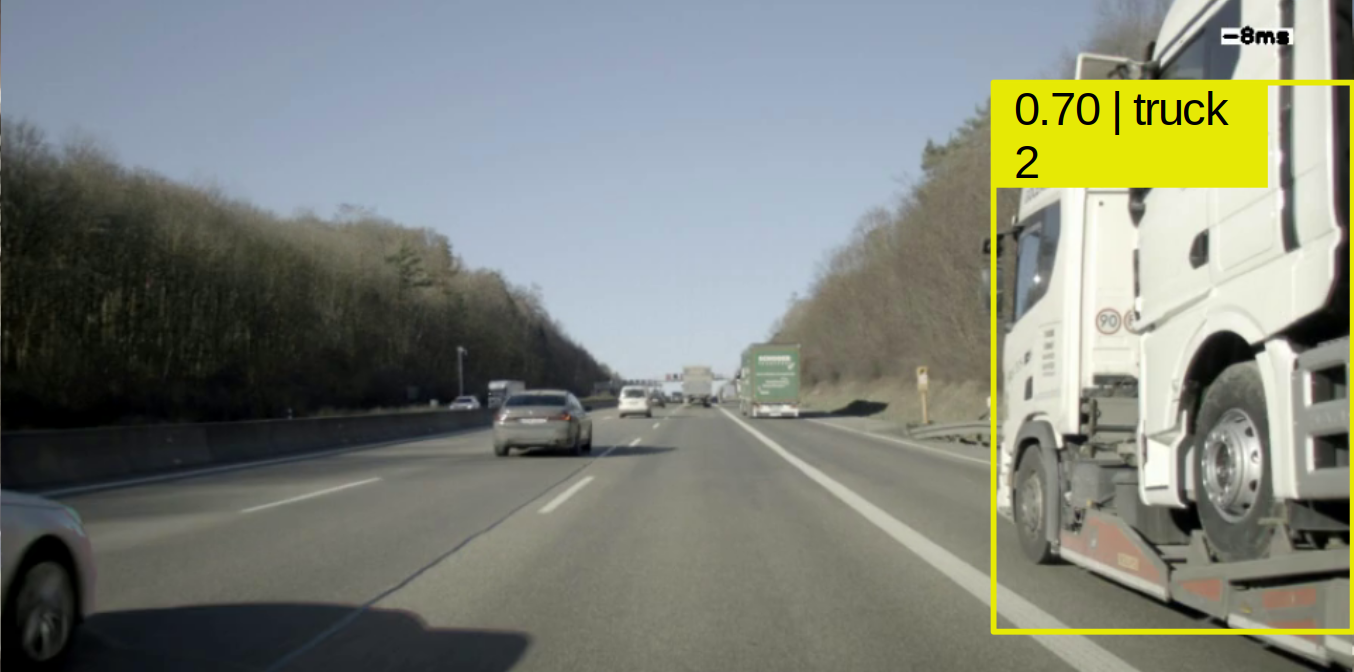}
    \end{subfigure}
    \hfill
    \begin{subfigure}[b]{0.24\textwidth}
        \centering
        \includegraphics[trim=0 0 0 0,clip,width=\textwidth]{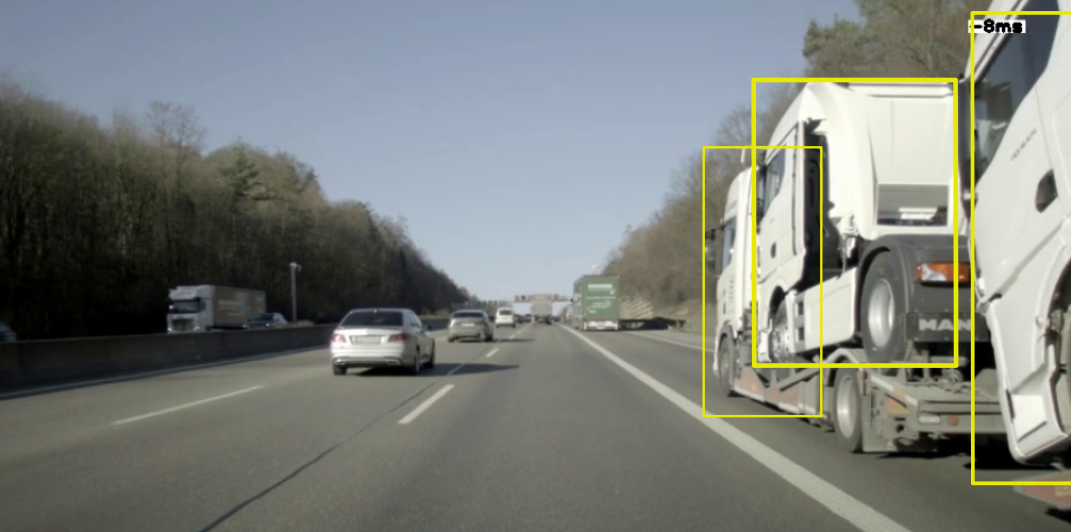}
    \end{subfigure}
    \hfill
    \begin{subfigure}[b]{0.24\textwidth}
        \centering
        \includegraphics[trim=0 0 0 0,clip,width=\textwidth]{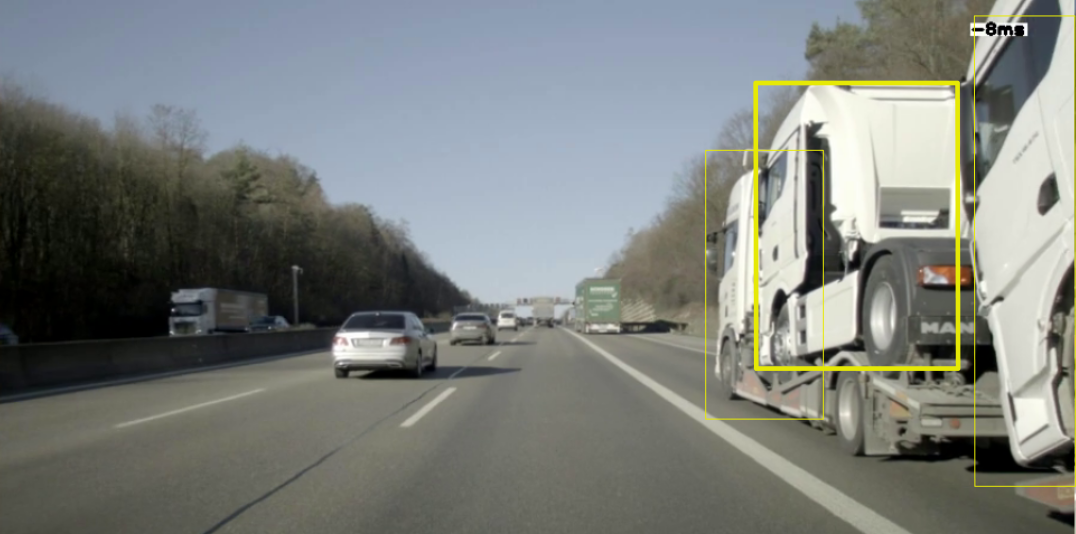}
    \end{subfigure}
    \hfill
    \begin{subfigure}[b]{0.24\textwidth}
        \centering
        \includegraphics[trim=0 0 0 0,clip,width=\textwidth]{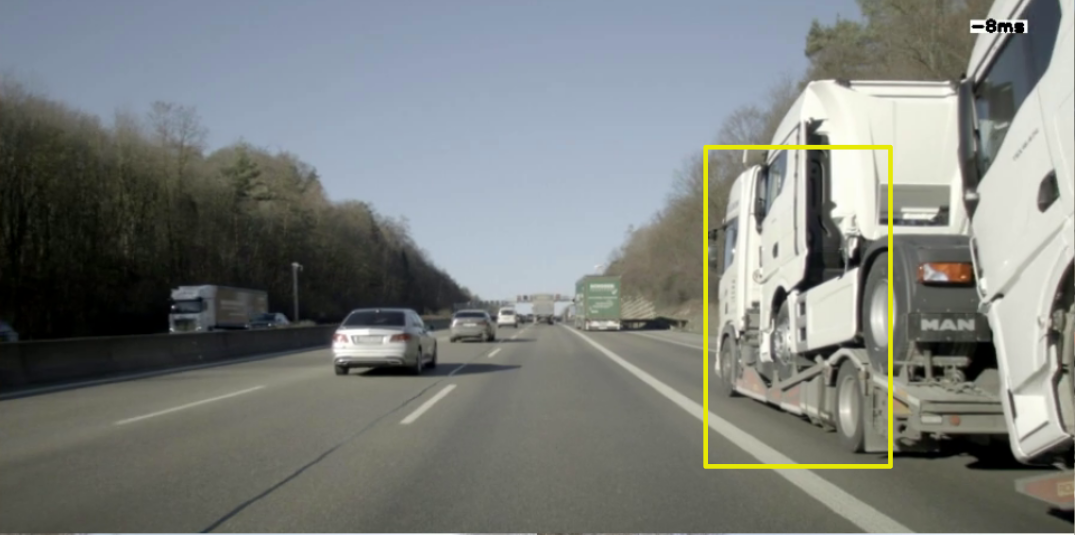}
    \end{subfigure}
    
    \vspace{0.2cm}
    
    \begin{subfigure}[b]{0.24\textwidth}
        \centering
        \includegraphics[trim=0 0 0 0,clip,width=\textwidth]{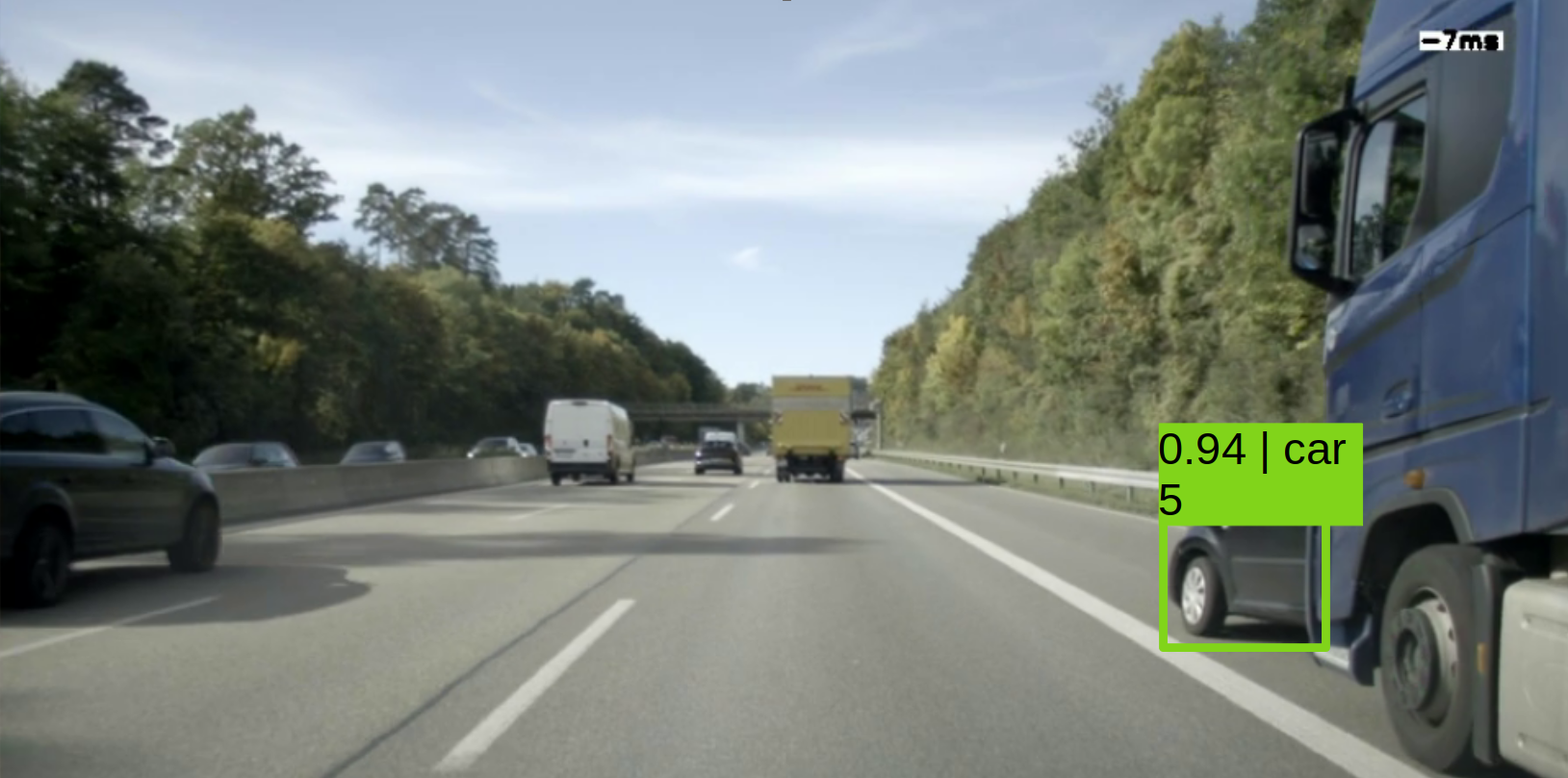}
        \caption{Reference frame}
    \end{subfigure}
    \hfill
    \begin{subfigure}[b]{0.24\textwidth}
        \centering
        \includegraphics[trim=0 0 0 0,clip,width=\textwidth]{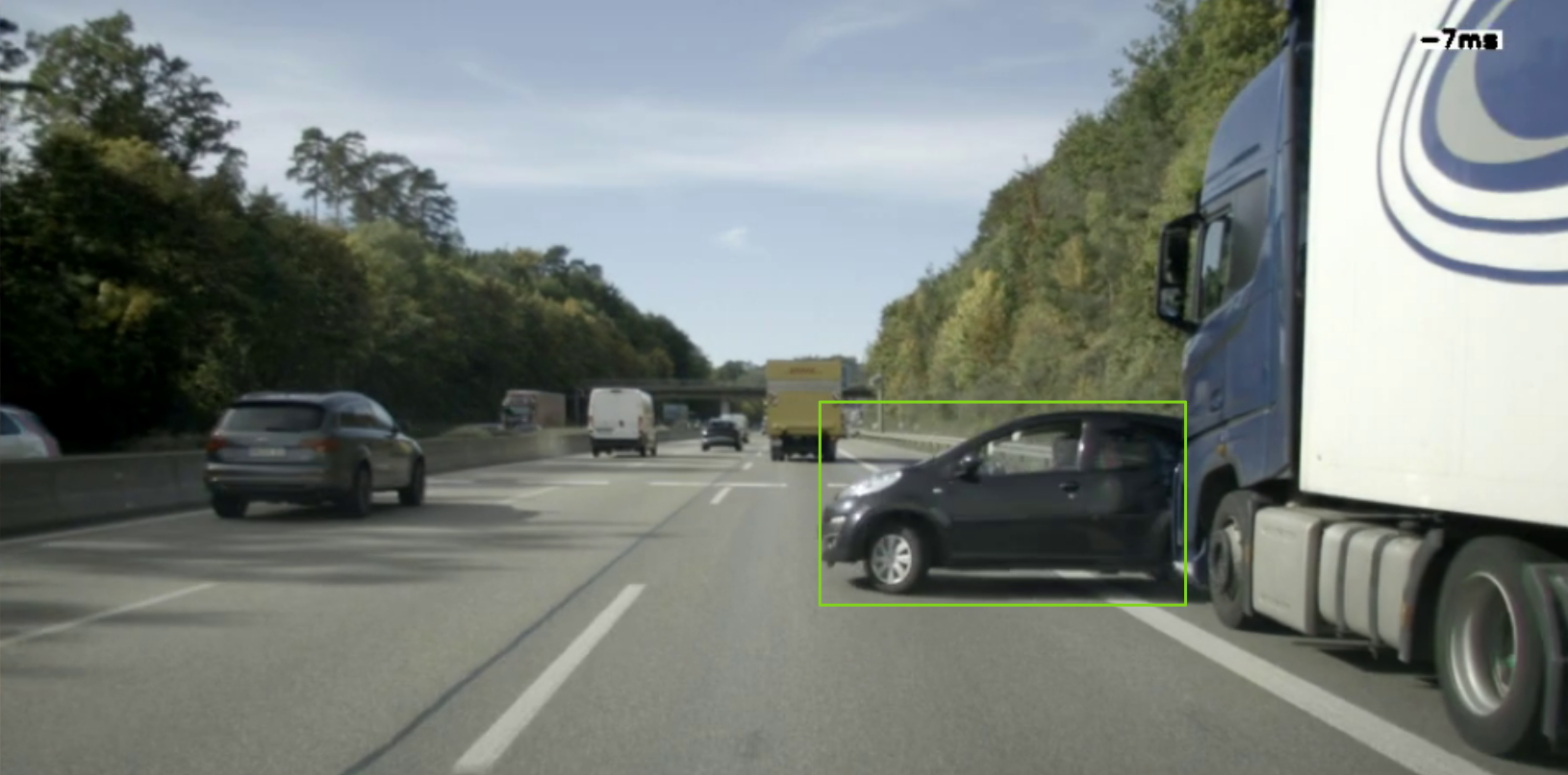}
        \caption{Bbox IoU}
    \end{subfigure}
    \hfill
    \begin{subfigure}[b]{0.24\textwidth}
        \centering
        \includegraphics[trim=0 0 0 0,clip,width=\textwidth]{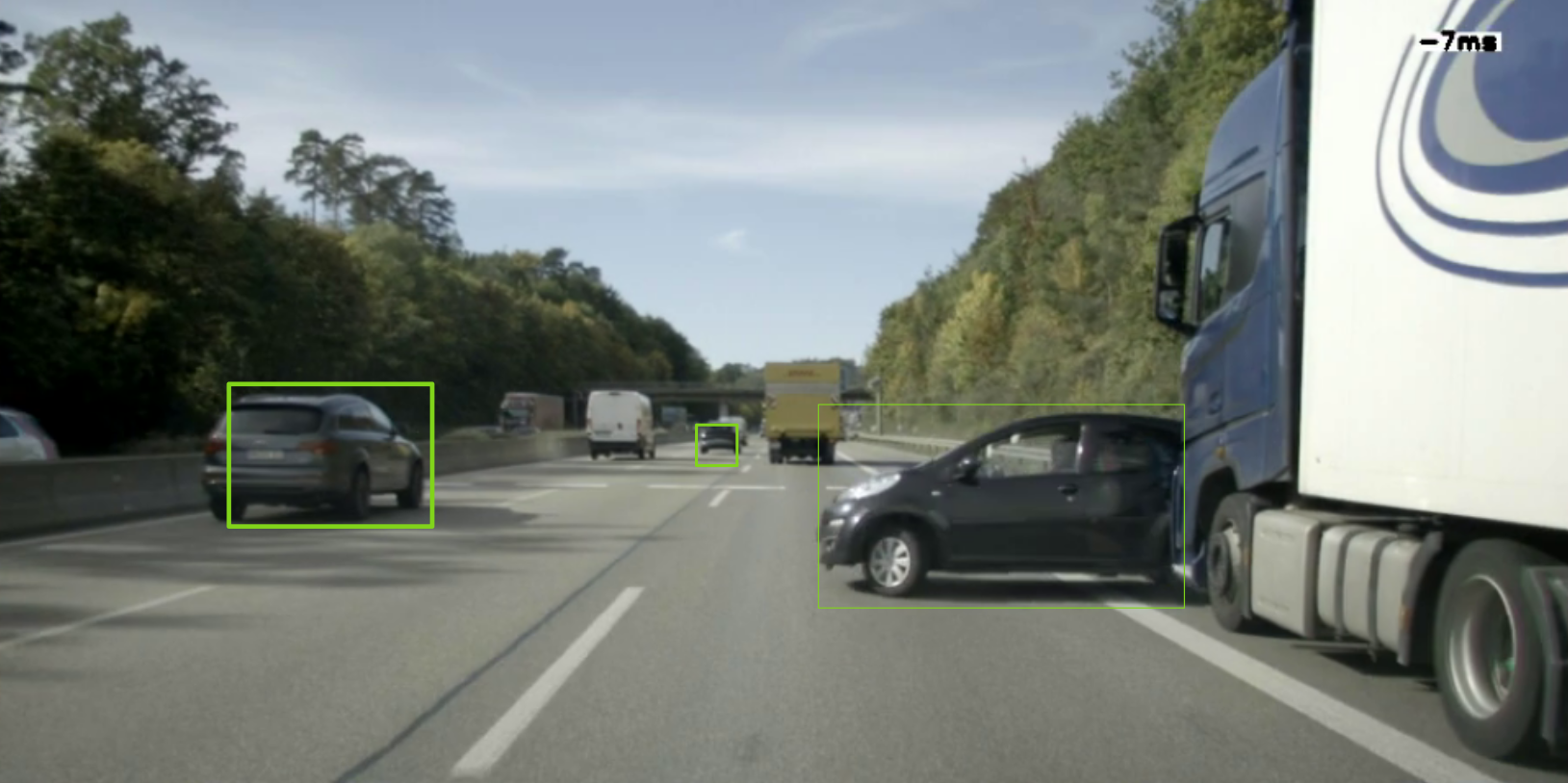}
        \caption{SwaV ReID}
    \end{subfigure}
    \hfill
    \begin{subfigure}[b]{0.24\textwidth}
        \centering
        \includegraphics[trim=0 0 0 0,clip,width=\textwidth]{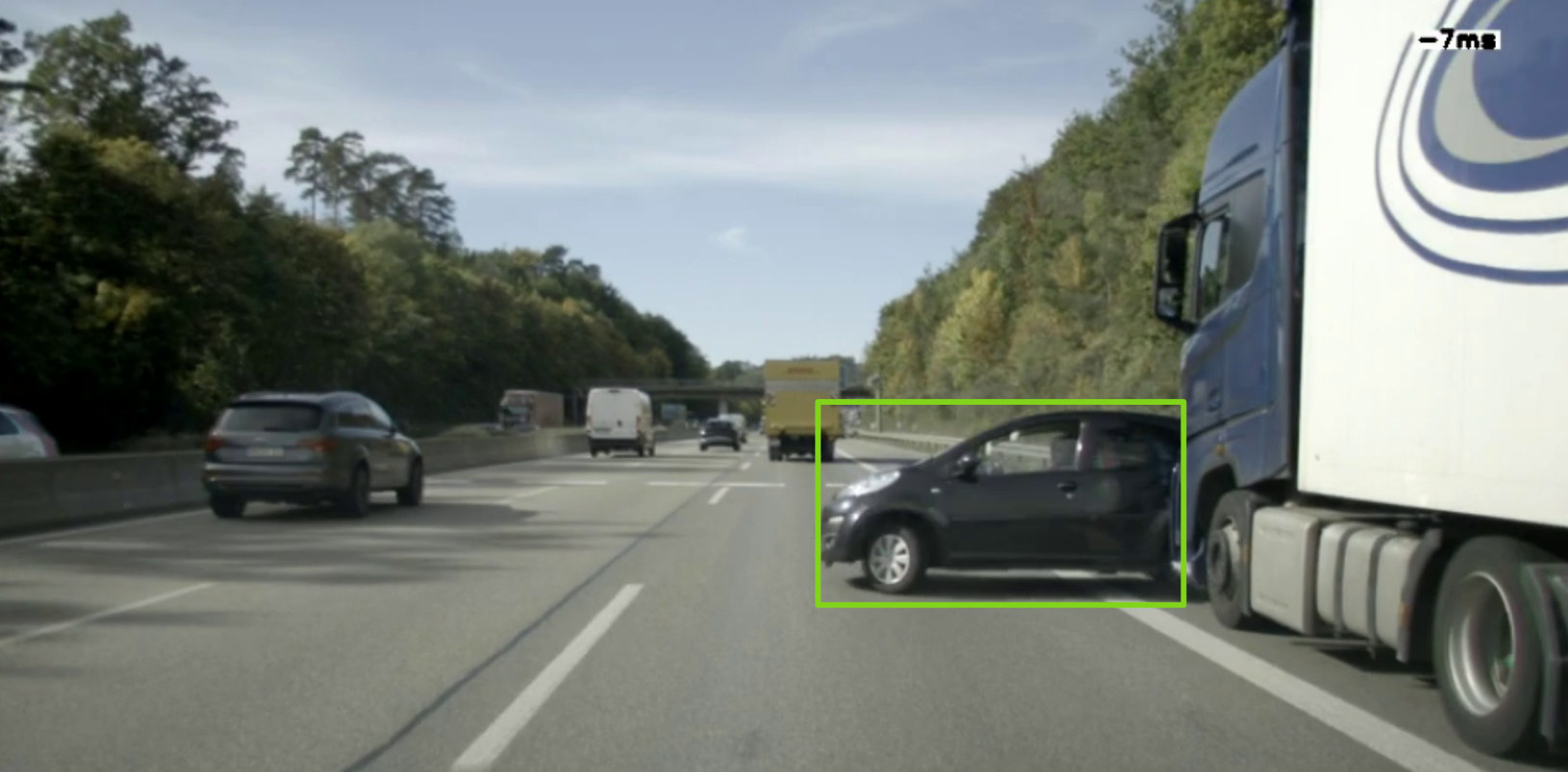}
        \caption{SubCo ReID (Ours)}
    \end{subfigure}



    \caption{Qualitative comparison of association methods for driving scenes captured by out-of-distribution on a German highway. The setup and model parameters are equivalent to Figure 3.
    }
    \label{fig:qualitative-example-mot-sclass}
\end{figure*}

On the \gls{bdd} dataset, the configuration using purely \gls{reid} similarity-based matching achieves the highest \gls{mota} of $42.3\%$ as well as the lowest number of ID switches. Surprisingly, the configuration using \gls{reid} in the second matching stage outperforms \gls{iou} for both settings in the first stage. This result differs from ~\cite{zhang2022bytetrack} experiments, which observed \gls{iou}-based matching outperforming \gls{reid} similarities for low-confidence detections.
They attributed the low confidence to occlusion or motion blur, which they linked to \gls{reid} features becoming unreliable.
We did not observe this effect using \gls{reid} features learned by our \gls{subco} method. This could indicate that the learned representations become more robust against disturbances like partial occlusions by observing them in the training stage.

In \tabref{tab:ablation_trackers}, we report results for using the \gls{subco} \gls{reid} features with varying combinations of base detectors and tracking logic.
First, we replace the BYTE tracker with the DeepSORT tracker~\cite{wojke2017simple}. The resulting performance drops by $-12.4\%$ for MOTA on the MOT17 dataset and $-5.7\%$ for mMOTA on the \gls{bdd} datasets. However, the \gls{subco} \gls{reid} features boost the performance of the supervised DeepSORT tracker by $+0.8\%$ for MOTA on the MOT17 benchmark.
Second, we replace the YOLOX base detector with CenterNet~\cite{zhou2020centertrack} initialized from the checkpoint provided  by FairMOT~\cite{zhang2021fairmot}. Please note that the FairMOT~\cite{zhang2021fairmot} tracker is integrated and trained jointly with the base detector and requires detection annotations. Therefore, we compare the tracking performance with a BYTE tracker using our learned \gls{reid} features. While the slightly inferior accuracy of the CenterNet detector dampens the MOTA on MOT17 by $-2.4\%$ compared to the YOLOX configuration, the IDF1 score only decreases slightly by $-0.7\%$, possibly due to missed detections by the detector.

\subsection{Qualitative Evaluation}

In \figref{fig:qualitative-example-mot} and \figref{fig:qualitative-example-mot-sclass}, we compare association strengths based on bounding box \gls{iou} of propagated bounding boxes~\cite{bewley2016simple,zhang2022bytetrack}, SwaV features~\cite{caron2020swav} similarities, and \gls{reid} feature similarities learned by our proposed self-supervised training strategy.
The demo sequences depict frame pairs that caused failure cases of the \gls{iou}-based matching strategy and the  SwaV~\cite{caron2020swav} \gls{reid} features, in which our learned \gls{reid} features generate accurate and high confidence associations. 
The failure cases of \gls{iou}-based matching in (b) are caused by close-by trajectories of instances of the same object type. Further, the SwaV~\cite{caron2020swav} features also fail to discriminate among pedestrian or vehicle instances, presumably due to the cluster assignment strategy used for training the model parameters.
On the bottom row, we show an example of a poorly illuminated scene, resulting in lower confidence detection scores. Due to the directed light source by the headlights of the ego-car, the illumination of the moving bicycle and rider changes noticeably. As the \gls{iou}-based matching implemented in BYTE~\cite{zhang2022bytetrack} is weighted by the detection confidence, the track of the rider is lost in (b). Similarly, the SwaV features depicted in (c) result in low association scores for the rider between the two frames.
This suggests that the SwaV feature representations differ noticeably under these illumination variations. In contrast, the \gls{subco} \gls{reid} feature representations appear robust against illumination variation and poor lighting conditions and yield a confident re-identification of the rider instance.

\section{Conclusion}
In this work, we proposed a self-supervised learning approach tailored for automated driving domains that enables training a multi-object tracker on frame sequence resulting in temporally consistent \gls{reid} representations for objects.
Extensive evaluations on the MOT17 and \gls{bdd} multi-object tracking benchmarks demonstrate that \gls{reid} models trained on our proposed \gls{subco} loss learns expressive and robust \gls{reid} feature representations that achieve a mean HOTA score of $43.3\%$ across all the class types for \gls{mot} on the \gls{bdd} benchmark, thereby setting the new state of the art for self-supervised MOT and perform on par with fully supervised learning methods.

\noindent \textit{Discussion of limitations and future work:} Our initial experiments study the effectiveness of our approach, even though the frame sequences $<2s$ are relatively short compared to the duration of driving or human activities. Extending the sequence length requires finer tuning of hyperparameters such as step size, using gradient accumulation to split the sequence over multiple GPUs, or defining the \textit{SubCo} loss over multiple step sizes at once. Further, we analyzed our methods only on standalone \gls{reid} models, while our method can also be applicable to train a tracking head for joint tracking and detection architectures. 










\footnotesize
\bibliographystyle{IEEEtran}
\bibliography{egbib}

\end{document}